\documentclass{article} 
\usepackage{iclr2027_conference,times}

\usepackage{amsmath,amsfonts,bm}

\def\eqref#1{equation~\ref{#1}}

\def\1{\bm{1}}

\DeclareMathAlphabet{\mathsfit}{\encodingdefault}{\sfdefault}{m}{sl}
\SetMathAlphabet{\mathsfit}{bold}{\encodingdefault}{\sfdefault}{bx}{n}

\usepackage{hyperref}
\usepackage{url}
\usepackage[ruled,vlined,linesnumbered]{algorithm2e}
\usepackage{multirow}
\usepackage{adjustbox}
\usepackage{graphicx}
\usepackage{enumitem}
\usepackage[table]{xcolor}
\usepackage{booktabs}
\usepackage{natbib}
\usepackage{placeins}
\usepackage{float}
\usepackage{balance}
\usepackage{amsmath,amssymb,amsfonts}
\usepackage{needspace}
\usepackage{algorithmic}
\usepackage{graphicx}
\usepackage{wrapfig}
\usepackage{textcomp}
\usepackage{xcolor}
\usepackage{makecell}
\usepackage[caption=false,font=footnotesize]{subfig}

\title{Boosting Knowledge Graph
Foundation Models via Enhanced
Negative Sampling}

\author{Yinan Liu, Wenjin Xu, Zhiyuan Zha, Xiaochun Yang\thanks{Corresponding author}, Bin Wang}

\iclrfinalcopy 
\begin{document}

\maketitle

\begin{abstract}
Knowledge graphs (KGs) are the core backbone of numerous downstream tasks such as question answering and recommender systems. However, despite all this, KGs are often very incomplete. To perform zero-shot link prediction in unseen KGs, which have different entity/relation vocabularies from those used for pre-training, KG foundation models (KGFMs) receive a wide range of attention. Existing KGFMs often perform training using random negatives, which are constructed by replacing the head or tail entity of a positive triple with a random entity. However, these negatives are often constructed with limited quality, providing weak supervision for KGFM training. In this paper, we propose a simple yet effective negative sampling approach, KMAS, to enhance existing KGFMs. KMAS constructs hard negative triples through the updated relation embeddings generated from the existing KGFM's relation encoder. To further align with the KGFM’s evolving capability to distinguish between positive and negative samples during the training process, 
KMAS uses a heuristic strategy for the hard negative ratio adjustment throughout the whole training process: after a warmup phase, it increases the ratio linearly and then decreases linearly. Extensive experiments are conducted over 44 data sets. Experimental results demonstrate that KMAS can enhance many SOTA KGFMs without requiring excessive additional memory consumption, and it incurs modest additional time cost on most KGFMs.
\end{abstract}
\section{Introduction}
Knowledge graphs (KGs), which consist of massive amounts of knowledge in the form of triples (head entity, relation, tail entity), have become the core backbone of numerous downstream tasks such as question answering \citep{zhao2024kgcot} and recommender systems \citep{KGAT19}. However, despite all this, KGs are often very incomplete \citep{Pan2024Unifying, liu2026joint, liu2021joint, shen2018predicting}, making the task of link prediction \citep{sun2025vldb, Rossi2022Kelpie}
increasingly important. Traditional transductive link prediction methods (e.g., TransE \citep{brodes2013transe} and RotatE \citep{sun2019rotate}) learn entity/relation embeddings constrained by specific entity and relation vocabularies, lacking the ability to generalize to unseen entities/relations on KGs. To address the limitation, partially inductive link prediction approaches (e.g., \citep{liu2021indigo}) relax the constraint on entities and generalize to unseen entities while keeping the relation vocabulary fixed. Fully inductive link prediction approaches (e.g., \citep{lee2023ingram}) further relax the constraint on relations and generalize to both unseen entities and unseen relations, which motivates the development of knowledge graph foundation models (KGFMs). KGFMs learn transferable invariances across KGs \citep{galkin2023ultra} by pre-training on multiple KGs of different entity/relation vocabularies, thereby enabling zero-shot reasoning on unseen KGs.

Existing KGFMs \citep{galkin2023ultra,huang2025how,arun2025semmasemanticawareknowledge} usually adopt a common procedure: (1) construct a relation graph based on the KG; (2) apply a relation encoder on this relation graph to obtain relation embeddings; (3) use the obtained relation embeddings and employ an entity encoder on the KG to obtain final link prediction results. 
It is worth noting that all the training processes of these KGFMs' relation and entity encoders are based on random negative triples, which are constructed by replacing the head or tail entity of a positive triple with a random entity.  
%
Although straightforward and efficient, these negative triples are often constructed with limited quality 
(i.e., ``easy negative triples''), providing weak supervision. Since these easy negative triples do not dynamically adjust throughout the whole training process, they fail to align with the KGFM's evolving capability to distinguish between positive and negative samples.
For instance, given a positive triple (\textit{Obama}, \textit{born\_in}, \textit{Hawaii}), a randomly constructed negative triple like (\textit{Obama}, \textit{born\_in}, \textit{Basketball}) is trivial for a well-trained KGFM. Conversely, constructing ``hard negative triples'' such as (\textit{Obama}, \textit{born\_in}, \textit{Chicago})---which is factually incorrect yet semantically plausible---can provide more useful supervision signals to guide the KGFM to learn the transferable invariances across KGs, enhancing the performance of the KGFM. In this paper, our task is to construct such hard negative samples to enhance KGFMs.


It is found that many negative sampling methods \citep{wang2022simkgc, che2025hard, ko2025subgraph, li2026kgc, ding2026arns} have been developed for the task of KG link prediction. They are usually designed for the transductive setting \citep{che2025hard, ko2025subgraph, li2026kgc} or both the transductive and partially inductive settings \citep{wang2022simkgc}, and rarely for the fully inductive setting \citep{ding2026arns}. Since most of these negative sampling methods are trained and evaluated on a single KG, they cannot construct meaningful negatives for KGFMs to learn transferable invariances across KGs.
Thus, when they are directly leveraged to enhance existing KGFMs, they usually cannot achieve good performance, which has been verified in our experiments.

To enhance existing \underline{\textbf{K}}GF\underline{\textbf{M}}s, we propose a simple yet effective neg\underline{\textbf{A}}tive \underline{\textbf{S}}ampling method KMAS. 
To construct hard negative triples for improving the training process of the KGFM, KMAS leverages the relation embeddings generated by the relation encoder of an existing KGFM to construct the tail (head) entity distribution, which enables hybrid negative sampling. 
In order to align with the KGFM's evolving capability to distinguish between positive and negative samples during the training process, KMAS iteratively updates relation embeddings using the encoder from the last training iteration. This update refines the tail (head) entity distribution and thereby generates progressively more effective negative samples. Notably, KMAS dynamically adjusts the ratio of hard negative triples throughout the training process: after a warmup phase that employs pure random negative sampling (as in existing KGFMs), it linearly increases the ratio of hard negative triples to a peak, then linearly decreases it in subsequent iterations.

The main contributions of this paper are summarized as follows:
We conduct the first investigation of negative sampling for KGFMs. We propose a simple yet effective negative sampling method by constructing hard negative samples with a heuristic hard negative ratio adjustment strategy, which is flexible to adapt to many KGFMs. We conduct extensive experiments on $44$ data sets. Experimental results demonstrate that our proposed negative sampling method can improve the performance of many SOTA KGFMs without requiring excessive additional memory consumption, and it incurs modest additional time cost on most KGFMs.

\section{Preliminaries}
A knowledge graph (KG) is denoted by $\mathcal{G} = (\mathcal{V}, \mathcal{R}, \mathcal{T})$, where $\mathcal{V}$ is a set of entities, $\mathcal{R}$ is a set of relations, and $\mathcal{T} \subseteq \mathcal{V} \times \mathcal{R} \times \mathcal{V}$ is a set of triples. A triple $(h, r, t) \in \mathcal{T}$ connects the head entity $h \in \mathcal{V}$ with the tail entity $t \in \mathcal{V}$ via the relation $r \in \mathcal{R}$.
Given a KG $\mathcal{G} = (\mathcal{V}, \mathcal{R}, \mathcal{T})$, the corresponding relation graph $\mathcal{G}_{\mathcal{R}} = (\mathcal{V}_{\mathcal{R}}, \mathcal{E}_{\mathcal{R}})$ is a directed graph where nodes $\mathcal{V}_{\mathcal{R}}$ are the relations $\mathcal{R}$. The edges $\mathcal{E}_{\mathcal{R}}$ capture interactions between relations based on their connectivity in $\mathcal{T}$. Given a KG $\mathcal{G} = (\mathcal{V}, \mathcal{R}, \mathcal{T})$, the link prediction task aims to infer missing facts based on observed ones. Formally, given a query $(h, r, ?)$ (resp. $(?, r, t)$), the goal is to predict the missing tail entity $t \in \mathcal{V}$ (resp. head entity $h \in \mathcal{V}$).
During inference, for a query $(h, r, ?)$, we calculate scores for all candidate entities $t' \in \mathcal{V}$ and rank them in descending order. The objective is to rank the ground-truth entity higher than other candidates.
We denote $\mathcal{G}_{train} = (\mathcal{V}_{train}, \mathcal{R}_{train}, \mathcal{T}_{train})$ as the training graph and $\mathcal{G}_{test} = (\mathcal{V}_{test}, \mathcal{R}_{test}, \mathcal{T}_{test})$ as the inference graph. 
KG foundation models (KGFMs) are usually trained on multiple KGs to learn transferable invariances across KGs and tested on entirely distinct KGs. 
As the entities and relations in the training and testing KGs are totally disjoint, KGFMs can perform the zero-shot link prediction, which is equivalent to the fully inductive setting.

For existing KGFMs, we use $\text{Encoder}_{\theta_r}$ to denote the relation encoder that generates relation embeddings. $\text{Encoder}_{\theta_e}$ denotes the entity encoder that generates entity embeddings. $f_{\omega}$ denotes an MLP that maps these entity embeddings to final scores.
$\theta_r$ denotes the parameters of the GNN architecture of $\text{Encoder}_{\theta_r}$. $\theta_e$ denotes the parameters of the GNN architecture of $\text{Encoder}_{\theta_e}$. 
$\omega$ denotes the parameters of the MLP that generates the final scores. For KGFMs, the GNN architecture often adopts NBFNet~\citep{zhu2021neural}. In KGFMs, a labeling trick is used to initialize relations to generate conditional relation embeddings based on the relation graph $\mathcal{G}_{\mathcal{R}}$, which means that $r$ of the given query (i.e., $(h, r, ?)$ or $(?,r,t)$) is initialized as an all-one vector, while other relations are initialized as all-zero vectors. $\text{Encoder}_{\theta_r}$ uses initialized relation embeddings to generate conditional relation embeddings for $\text{Encoder}_{\theta_e}$. 
In $\text{Encoder}_{\theta_e}$, the entity in query is initialized as the conditional relation embedding of $r$, and other entities are initialized as all-zero vectors. Then, $\text{Encoder}_{\theta_e}$ uses the initialized entity embeddings and conditional relation embeddings to 
generate entity embeddings for $f_{\omega}$ to obtain the scores, which are used to predict the missing entity in the given query.

\begin{figure*}[t] 
    \centering
    \includegraphics[width=0.85\textwidth]{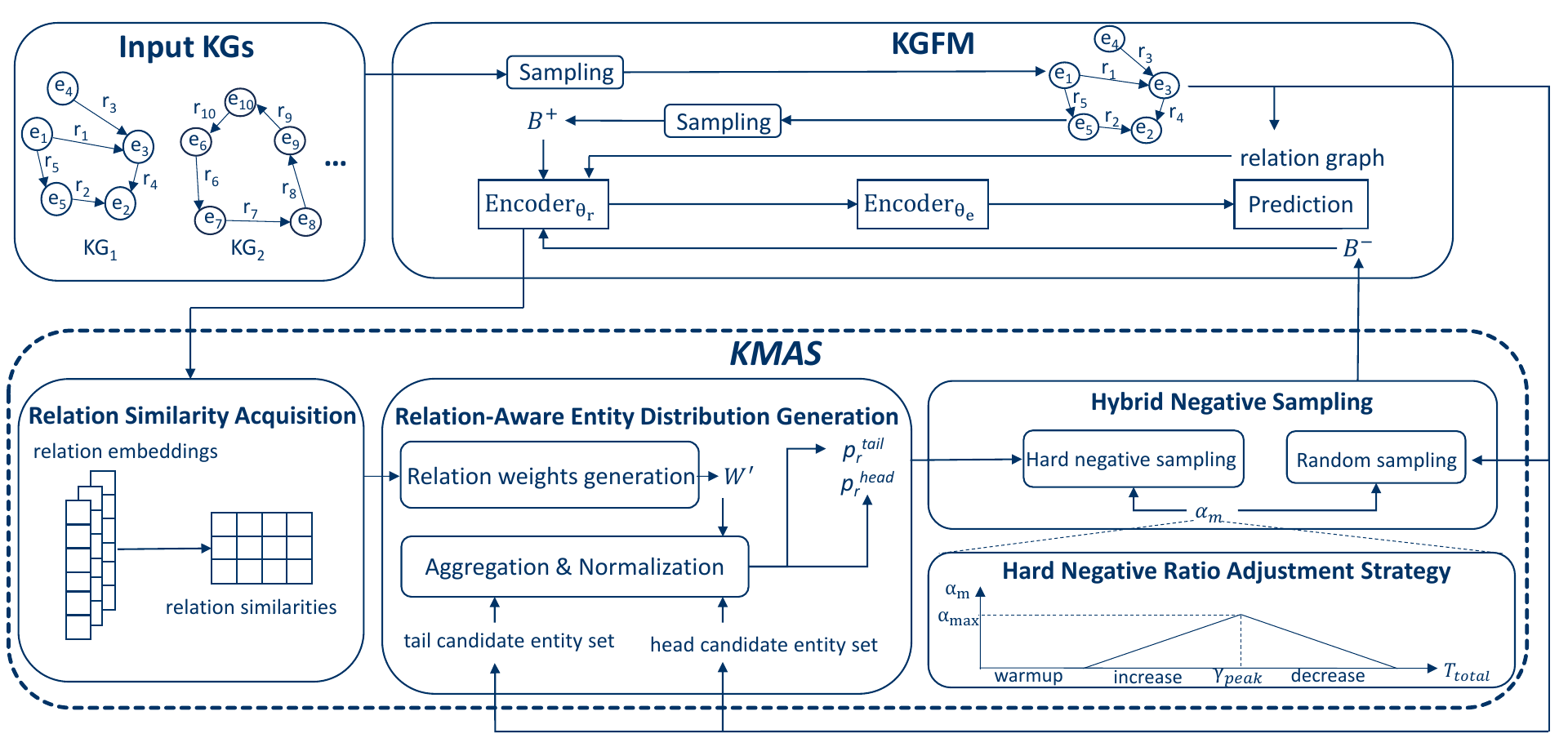}
    \caption{The overall training process of KMAS.} 
    \label{fig:framework}
\end{figure*}

\section{The Method KMAS}


The overall method, KMAS, is shown in Figure~\ref{fig:framework}. Here, we briefly introduce it. Specifically, during each iteration of an existing KGFM's training process, KMAS first calculates relation similarities to identify the relations that are semantically similar to the target relation of each positive triple based on the relation embeddings generated through $\text{Encoder}_{\theta_r}$ based on the relation graph (Section \ref{sec:sm_acquire}). Next, using these similarities, KMAS assigns weights to entities to construct head (tail) entity distributions for hard negative sample generation (Section \ref{sec:neg_gen}). Then, a hybrid sampling strategy is employed to construct the final negative triple set for each triple by combining random and hard negative samples according to different distributions (Section \ref{sec:hybrid}). Furthermore, KMAS dynamically adjusts the ratio of hard negative samples versus random negative samples throughout training iterations, which follows a trade-off policy by combining both ``easier first" as curriculum learning and ``harder first" as hard example mining: beginning with a warmup phase without hard negative samples, increasing linearly to a peak, and then decreasing linearly (Section \ref{sec:hardRatio}). Thus, the KGFM's $\text{Encoder}_{\theta_r}$ can be trained based on the training set consisting of positive triples and constructed negative triples (Section \ref{sec:trainAndInfer}). After multiple iterations, the relation embeddings generated from $\text{Encoder}_{\theta_r}$, together with the KGFM's entity encoder $\text{Encoder}_{\theta_e}$ and $f_{\omega}$, are utilized to score all candidates for each test triple (Section \ref{sec:trainAndInfer}).

\subsection{Relation Similarity Acquisition}
\label{sec:sm_acquire}

To construct high-quality hard negative samples for positive triples, we first utilize the relation embeddings generated by $\text{Encoder}_{\theta_r}$ to measure the similarities between the relations. For the relation $r$ of a given positive triple $(h, r, t)$, we input it into $\text{Encoder}_{\theta_r}$ to obtain relation embeddings for all relations in ${\mathcal{G}}$ as follows:
$H \leftarrow \text{Encoder}_{\theta_r}(\mathcal{G}_\mathcal{R}, r, r_t), 
$
where $d$ denotes the dimension of embeddings in $\text{Encoder}_{\theta_r}$, $H \in \mathbb{R}^{|\mathcal{R}| \times d}$ denotes the set of relation embeddings. To obtain sufficiently informative relation embeddings, we leverage those produced by $\text{Encoder}_{\theta_r}$ after a warmup training phase, as detailed in Section \ref{sec:hardRatio}. If SEMMA's relation encoder is selected as $\text{Encoder}_{\theta_r}$, $r_t$ is set to the text of relations. For other KGFMs (i.e., ULTRA and MOTIF), $r_t$ is set to none. 
Then, based on $H$, we can calculate the similarity 
vector $\mathbf{S} \in \mathbb{R}^{|\mathcal{R}|}$ between $r$ and the relation $r_o$ in $\mathcal{R}$ as follows:
$\mathbf{S}({r, r_o}) = \text{Sim}(\mathbf{h}_{r}, \mathbf{h}_{r_o})
$,
where the function Sim($\cdot$) is implemented by the cosine similarity, $\mathbf{h}_{r}$ denotes the relation embedding of $r$ and $\mathbf{h}_{r_o}$ denotes the relation embedding of $r_o$.
Specifically, we set  $\mathbf{S}({r, r})$ to $-\infty$, which will prevent the relation itself from sampling in the following process. Based on this, we construct the vector $W_{r} \in \mathbb{R}^{|\mathcal{R}|}$ for $r$ to store relation similarities using the Softmax function with a temperature coefficient $\tau$ as follows:
\begin{equation} \label{eq:weight_calc}
W_{r,j} = \frac{\exp(\mathbf{S}_{r, j} / \tau)}{\sum^{|\mathcal{R}|}_{j' = 1} \exp(\mathbf{S}_{r, j'} / \tau)}, j=1,2, \ldots,\mathcal{|R|}.
\end{equation}

\subsection{Relation-Aware Entity Distribution Generation} \label{sec:neg_gen}
First, we sort the weights of relations in $W_{r}$ in descending order to obtain $W^*_{r}$. Next, for $r$, we select the top $K$ relations $R_{topK}{(r)}$ whose sum of weights reaches or exceeds a threshold $q$ as follows:

\begin{equation}
    K = \underset{z}{\arg\min} \; z, \quad \text{s.t.} \quad \sum_{y=1}^{z} W^{*}_{r,y} \geq q, \quad z \in \{1, \ldots, |\mathcal{R}|\}.
\end{equation}
For the relations not in $R_{topK}{(r)}$, we set their weights to zero, which can ensure the generated negatives are not semantically odd. Subsequently, we reweight the selected relations to construct a weight matrix $W'$ by calculating the weight $W'_{r, r'}$ for every relation $r$ and its corresponding selected relation $r'$ as follows:
$W'_{r, r'} = W^*_{r, r'}/\sum_{r' = 1}^K W^*_{r, r'}, r'=1,2, \ldots, K.
$
To generate hard negative samples for the given positive triple $(h, r, t)$, for each selected relation $r' \in R_{topK}(r)$, we first construct its tail candidate entity set $w^{tail}_{r'}$ (resp. $w^{head}_{r'}$) by searching entities in $\mathcal{V}$ that can act as tail (resp. head) entities of $r'$ in $\mathcal{G}$.
Thus, we can obtain the weight $x^{tail}_{r, e}$ of an entity $e$ as the tail entity of $r$ by aggregating the relation weights as follows:

\begin{equation} \label{eq:tail_weight}
x^{tail}_{r, e} = \sum_{r' \in R_{topK}(r)} W'_{r,r'} \cdot I_{tail}(r', e),  \ \  I_{tail}(r', e) = 
        \begin{cases}
            1 & \text{if } e \in w^{tail}_{r'}   \\
            0 & \text{otherwise}.
        \end{cases}
\end{equation}
It can be seen that if an entity frequently appears as the tail entity of the relation that is similar to $r$, this entity will be assigned a high probability score. Then, we normalize these weights of entities to obtain a tail entity probability distribution $\mathcal{P}^{tail}_{r}$ for $r$ as follows:
$\mathcal{P}^{tail}_{r}(e) = x^{tail}_{r, e}/\sum_{e' \in \mathcal{V}} x^{tail}_{r, e'}.
$
Note that, we can obtain the weight $x^{head}_{r, e}$ of an entity $e$ as the head entity of $r$ and the head entity probability distribution $\mathcal{P}^{head}_{r}$ for $r$ in a similar way, which is omitted due to the length of the paper.

\subsection{Hybrid Negative Sampling} \label{sec:hybrid}
To improve the model's performance, we propose a hybrid sampling strategy to provide more challenging negative samples while still retaining the simpler samples, preserving the fundamental ability to distinguish between positive and negative samples. We denote $\mathcal{B}^+$ as a batch of positive triples. For each positive triple $(h, r, t)$ in $\mathcal{B}^+$, we construct its corresponding negative sample set $\mathcal{B}^{-}_{(h ,r, t)}$. Specifically, for $(h, r, t)$, we draw $N_{rand}$ entities
randomly from $\mathcal{V}$ to obtain $\mathcal{T}^{rand}_{(h, r, t)}$ (resp. $\mathcal{H}^{rand}_{(h, r, t)}$), and 
draw the $N_{hard}$ entities according to the 
distribution $\mathcal{P}^{tail}_r$ (resp. $\mathcal{P}^{head}_r$) to obtain $\mathcal{T}^{hard}_{(h, r, t)}$ (resp. $\mathcal{H}^{hard}_{(h, r, t)}$). The number of random negative samples $N_{rand}$ and hard negative samples $N_{hard}$ are calculated as follows: 
$N_{rand} = \lfloor N \cdot (1 - \alpha_m) \rfloor, \ \ N_{hard} = N - N_{rand}.$
Note that $\alpha_m$ denotes the current hard negative ratio, which will be calculated in Section \ref{sec:hardRatio}.
The negative sample sets $\mathcal{C}^{tail}_{(h, r, t)}$ and $\mathcal{C}^{head}_{(h, r, t)}$ are constructed as follows:
\begin{equation}
\label{eq:negetive_samples_tail}
    \mathcal{C}^{tail}_{(h, r, t)} =
    \{(h, r, t') | t' \in
        \mathcal{T}^{rand}_{(h, r, t)} \cup \mathcal{T}^{hard}_{(h, r, t)}\,  \}, \ \  \mathcal{C}^{head}_{(h, r, t)} =\{(h', r, t) | h' \in
        \mathcal{H}^{rand}_{(h, r, t)} \cup \mathcal{H}^{hard}_{(h, r, t)}\,  \}
\end{equation}

Based on these, we can construct the final negative sample set $\mathcal{B}^-_{(h, r, t)}$ for $(h, r, t)$ as follows:
\begin{equation}
\label{eq:hard_negative_samplings}
    \mathcal{B}^-_{(h, r, t)} = 
    \begin{cases}
        \mathcal{C}^{tail}_{(h, r, t)}  & \text{if } (h, r, t) \text{ is in the first half of } \mathcal{B}^+, \\
        \mathcal{C}^{head}_{(h, r, t)}  & \text{otherwise}.     
    \end{cases}
\end{equation}

Thus, by constructing negative samples for each positive triple in $\mathcal{B}^+$ according to the above procedure, we can obtain the corresponding negative sample set $\mathcal{B}^-$ of $\mathcal{B}^+$.
Here, we use a rigorous negative masking mechanism, corresponding to the function of \texttt{strict\_negative\_mask} in our implementation, to eliminate the risk of generating negatives containing false negatives, aligning with existing KGFMs. For any given query $(h, r, ?)$, this mechanism explicitly searches the entire set of triples to identify all tail entities associated with the same head entity and relation, subsequently excluding them during the negative sampling process.


\label{sec:dynamic_scheduling}

\subsection{Hard Negative Ratio Adjustment Strategy} 
\label{sec:hardRatio}
Inspired by ~\citep{wang2021survey} combining ``easier first" as curriculum learning and ``harder first" as hard example mining, we propose a strategy for hard negative sampling to dynamically adjust the ratio between random negative samples and our constructed hard negative samples. We define the hard negative ratio $\alpha_m$ as a 
piecewise linear function, which is controlled by 
the warmup ratio $\beta_{warm}$ ranging from 0 to 1, the 
peak position $\gamma_{peak}$, and the maximum ratio of the hard negative samples $\alpha_{max}$. Crucially, $\gamma_{peak}$ determines the point where $\alpha_m$ reaches its 
maximum value $\alpha_{max}$ within the remaining training 
process after the warmup phase, which means $\gamma_{peak}$ is greater than $\beta_{warm}$ and less than $1$. Let $m$ denote the $m$-th iteration of the whole training iterations and $T_{total}$ denote the total number of training iterations. 
We calculate the ratio of completed iterations $p_m$ as follows:
$
p_m = m/T_{total}.
$
We can calculate $\alpha_m$ for the current $m$-th iteration based on the following rules.
If $p_m$ is less than $\beta_{warm}$, we set $\alpha_m$ to $0$.
If $p_m$ is greater than or equal to $\beta_{warm}$ and less than $P_{peak}$, $\alpha_m$ is calculated as
    $\alpha_{max} \cdot (p_m - \beta_{warm})/(P_{peak} - \beta_{warm}).$ 
    If $p_m$ is greater than or equal to $P_{peak}$, $\alpha_m$ is calculated as
    $\alpha_{max} \cdot (1 - p_m)/(1 - P_{peak}),$
    where $P_{peak}$ equals $\beta_{warm} + \gamma_{peak}(1 - \beta_{warm})$.
For the hard negative ratio during the whole training process, this heuristic adjustment strategy is simple yet effective, which is verified by our experiments (Section \ref{sec:ablation}).

\subsection{Training}
\label{sec:trainAndInfer}
The KGFM is often trained by minimizing the binary cross-entropy loss function. 
We obtain the prediction score $s(h, r, t)$ of a triple $(h,r,t)$ via $f_\omega$. The optimization objective for a given positive triple $(h,r,t)$ and its corresponding negative triple set $\mathcal{B}^-_{(h, r, t)}$ is defined as follows:

\begin{equation} 
\begin{split}
\label{eq:bce_loss}
    \mathcal{L}(h,r,t) = -\log(\sigma(s(h,r,t)))- \sum_{(h', r, t') \in \mathcal{B}^-_{(h, r, t)}} w(h', r, t') \log(1 - \sigma(s(h', r, t'))),
\end{split}
\end{equation}
where $\sigma$ is the Sigmoid activation function. Additionally, considering that different negative samples have different importance to the training process of the KGFM, we assign the normalized weight to each negative sample using the Softmax function with a temperature coefficient ${\tau_{adv}}$ as follows:

\begin{equation} \label{eq:adv_weight}
w(h', r, t') = \frac{\exp({s(h', r, t')}/\tau_{adv})}{\sum_{(h_k, r, t_k) \in \mathcal{B}^-_{(h,r,t)} } \exp({s(h_k, r, t_k)}/\tau_{adv})}.
\end{equation}
Finally, the total optimization objective for the given batch $\mathcal{B}^+$ is calculated by averaging the losses over all positive triples in $\mathcal{B}^+$ as follows:

\begin{equation}
\label{eq:total_bce_loss}
    \mathcal{L}_{total} = \frac{1}{|\mathcal{B}^+|}{\sum_{(h, r, t) \in \mathcal{B}^+}{\mathcal{L}(h, r, t)}}.
\end{equation}

Notably, the processes of the relation similarity acquisition (Section \ref{sec:sm_acquire}) and the relation-aware entity distribution generation (Section \ref{sec:neg_gen}) depend on evolving relation embeddings.
Since the parameters of the relation encoder $\theta_r$ are updated continuously through the backpropagation mechanism during the training process of the KGFM, the embeddings of the relations also evolve throughout the whole training process. Consequently, even for the same KG and the same relation sampled in different iterations, the values of weights $W_r$ and the tail (resp. head) entity distribution $\mathcal{P}^{tail}_r$ (resp. $\mathcal{P}^{head}_r$) also change dynamically for each iteration. This mechanism ensures that the constructed negatives are time-variant. Although these negatives may not be the most difficult, they still represent the challenging negative samples for the current state of the KGFM. The details of the training process are shown in Algorithm 1 in Appendix \ref{appendix:algorithm}. Additionally, we adopt the same inference procedure as previous KGFM studies ~\citep{galkin2023ultra, huang2025how, arun2025semmasemanticawareknowledge}.

\subsection{Complexity Analysis}
\label{sec:comple}
We analyze KMAS's time and memory complexity (detailed in Appendix \ref{appendix:tiam_mem}).
$|\mathcal{V}|$ and $|\mathcal{R}|$ are the numbers of entities and relations, respectively. $|\mathcal{T}|$ is the number of triples in the KG. 
$d$ denotes the dimension of hidden embeddings. 
$L$ is the number of GNN layers. $\mathcal{|B^+|}$ is the batch size. $\bar{d}_{rel}$ is the average relation degree.
KMAS's time complexity is $\mathcal{O}( T_{total} \cdot \mathcal{|B^+|} \cdot (|\mathcal{R}|(d + \log |\mathcal{R}| + \bar{d}_{rel}) + |\mathcal{V}| + N))$. 
Since $|\mathcal{R}|$ is often small and $\mathcal{|B^+|}$/ $\bar{d}_{rel}$/$T_{total}$/$N$/$d$ are treated as constants, KMAS's time complexity is linear to $|\mathcal{V}|$. 
KMAS's memory cost is $\mathcal{O}( |\mathcal{B^+}| ( |\mathcal{R}| + |\mathcal{V}| +N ) )$.
Since $\mathcal{|R|}$ is usually small and $N$ is a constant, the memory cost is linear to $\mathcal{|B^+|} \cdot \mathcal{|V|}$. Base KGFMs need to store intermediate node embeddings across all $L$ layers to compute gradients during backpropagation, with a memory complexity of 
$\mathcal{O}(\mathcal{|B^+|} \cdot |\mathcal{V}| \cdot d \cdot L)$. Against $\mathcal{|B^+|} \cdot |\mathcal{V}| \cdot d \cdot L$, $\mathcal{|B^+|} \cdot \mathcal{|V|}$ is small, indicating a marginal overhead generated by KMAS, which is verified in Section ~\ref{sec:memeory_study}.

\section{Experiments}

\subsection{Experimental Setting}
\label{sec:experimental setting}
We use a unified pre-training corpus containing $3$ common KGs to train KGFMs: FB15k-237~\citep{fb15k237}, WN18RR~\citep{wn18rr}, and CoDEx-Medium~\citep{codex}. 
We perform experiments on 44 data sets categorized into 3 groups (detailed information of each data set is provided in Appendix \ref{appendix:datasets}): 
 Transductive data sets (9 KGs), including NELL-995 \citep{xiong2017deeppath}, CoDExSmall, CodDExLarge \citep{codex}, WDsinger, NELL23k, FB15k237(10), FB15k237(20), FB15k237(50) \citep{lv2020dynamic}, and Hetionet \citep{himmelstein2017systematic}.
     Partially inductive data sets (16 KGs), comprising 11 from GraIL \citep{teru2020inductive}, 2 from ILPC 2022 \citep{galkin2022open}, and 3 from INDIGO \citep{liu2021indigo}.
     Fully inductive data sets (19 KGs), consisting of 11 from INGRAM \citep{lee2023ingram} and 8 from MTDEA \citep{zhou2023multi}.
We use the same evaluation metrics, mean reciprocal rank (MRR) and Hits@10, as KGFM studies \citep{galkin2023ultra, huang2025how, arun2025semmasemanticawareknowledge}.
Both metrics are reported under the filtered ranking protocol \citep{brodes2013transe}.
 We perform KMAS on 5 KGFMs (i.e., ULTRA \citep{galkin2023ultra}, SCR \citep{wang2024towards}, MOTIF \citep{huang2025how}, SEMMA \citep{arun2025semmasemanticawareknowledge}, and FLOCK \citep{kim2025flock}), which are introduced in Appendix \ref{appendix:kgfms}.
We use the official implementations of these KGFMs.
$N$, $ q $, $\tau$, $\tau_{adv}$, $\beta_{\text{warm}}$, $\gamma_{\text{peak}}$,$\alpha_{\max}$ are set to $512$, $0.4$, $0.8$, $1$, $0.25$, $0.25$ and $0.4$ for ULTRA, MOTIF, SEMMA, FLOCK and $512$, $0.4$, $0.8$, $2$, $0.50$, $0.50$ and $0.10$ for SCR with KMAS as the negative sampling method, respectively. $T_{total}$ is set to $85090$ for ULTRA and SEMMA, 68067 for FLOCK, $10000$ for MOTIF and $9600$ for SCR.
All experiments are conducted on four NVIDIA 4080 GPUs with the same version of PyTorch/PyG. 

\setlength{\columnsep}{5pt}

\begin{wraptable}[13]{r}{0.43\textwidth}
\vspace{-\intextsep}
\vspace{-9pt}
\centering

\caption{Link prediction results.}
\label{tab:my_comparison}

\vspace{2pt}

\fontsize{8.5}{9}\selectfont
\setlength{\tabcolsep}{3pt}
\renewcommand{\arraystretch}{0.55}

\begin{adjustbox}{max width=\linewidth}
\begin{tabular}{ccc}
\toprule
\multirow{2}{*}{\bfseries\textit{Model (zero-shot)}}
& \multicolumn{2}{c}{\bfseries \textit {Total Avg (44 graphs)}} \\
\cmidrule(lr){2-3}
& \bfseries\textit{MRR}
& \bfseries\textit{Hits@10} \\
\midrule

ULTRA (ICLR'24)
& 0.360 & 0.524 \\
\rowcolor{blue!10}
ULTRA + KMAS
& \bfseries 0.367 & \bfseries 0.532 \\

\midrule

SCR (NIPS'25)
& 0.348 & 0.509 \\
\rowcolor{blue!10}
SCR + KMAS
& \bfseries 0.352 & \bfseries 0.513 \\

\midrule

MOTIF (ICML'25)
& 0.366 & 0.524 \\
\rowcolor{blue!10}
MOTIF + KMAS
& \bfseries 0.371 & \bfseries 0.533 \\

\midrule

SEMMA (EMNLP'25)
& 0.372 & 0.529 \\
\rowcolor{blue!10}
SEMMA + KMAS
& \bfseries 0.380 & \bfseries 0.539 \\

\midrule

FLOCK (ICLR'26)
& 0.380 & 0.551 \\
\rowcolor{blue!10}
FLOCK + KMAS
& \bfseries 0.386 & \bfseries 0.554 \\

\bottomrule
\end{tabular}
\end{adjustbox}

\vspace{-4pt}
\end{wraptable} 
\subsection{Effectiveness Study} \label{sec:effectiveness_etudy}
We utilize $\text{Encoder}_{\theta_r}$ to generate relation embeddings for the construction of hard negative samples in ULTRA, MOTIF, SEMMA. For FLOCK, we need to perform the whole forward process to obtain the relation embeddings, which are then used to construct the hard negative samples. In addition, for SCR, the relation graph is enhanced by deriving both global and query-aware relation embeddings and injecting them to guide semantic-conditioned message passing on the KG. Experimental results (Table \ref{tab:my_comparison}) show that KMAS can enhance the performance of the zero-shot link prediction task across various SOTA KGFMs over 44 data sets, demonstrating the effectiveness of KMAS. All KGFMs are reproduced via their open-source solutions. Detailed results on each data set are provided in Appendix \ref{appendix:full_res}. We further give a analysis of statistical significance and a comparison with model evolution and KMAS in Section \ref{sec:significance} to demonstrate the gain of KMAS is meaningful for the KGFM community.

\subsection{Efficiency Study} \label{sec:efficiency_study}
\begin{wrapfigure}{r}{0.55\textwidth}
    \vspace{-8pt}
    \centering
    \includegraphics[
    width=\linewidth,
    trim=0 20 0 0,
    clip
    ]{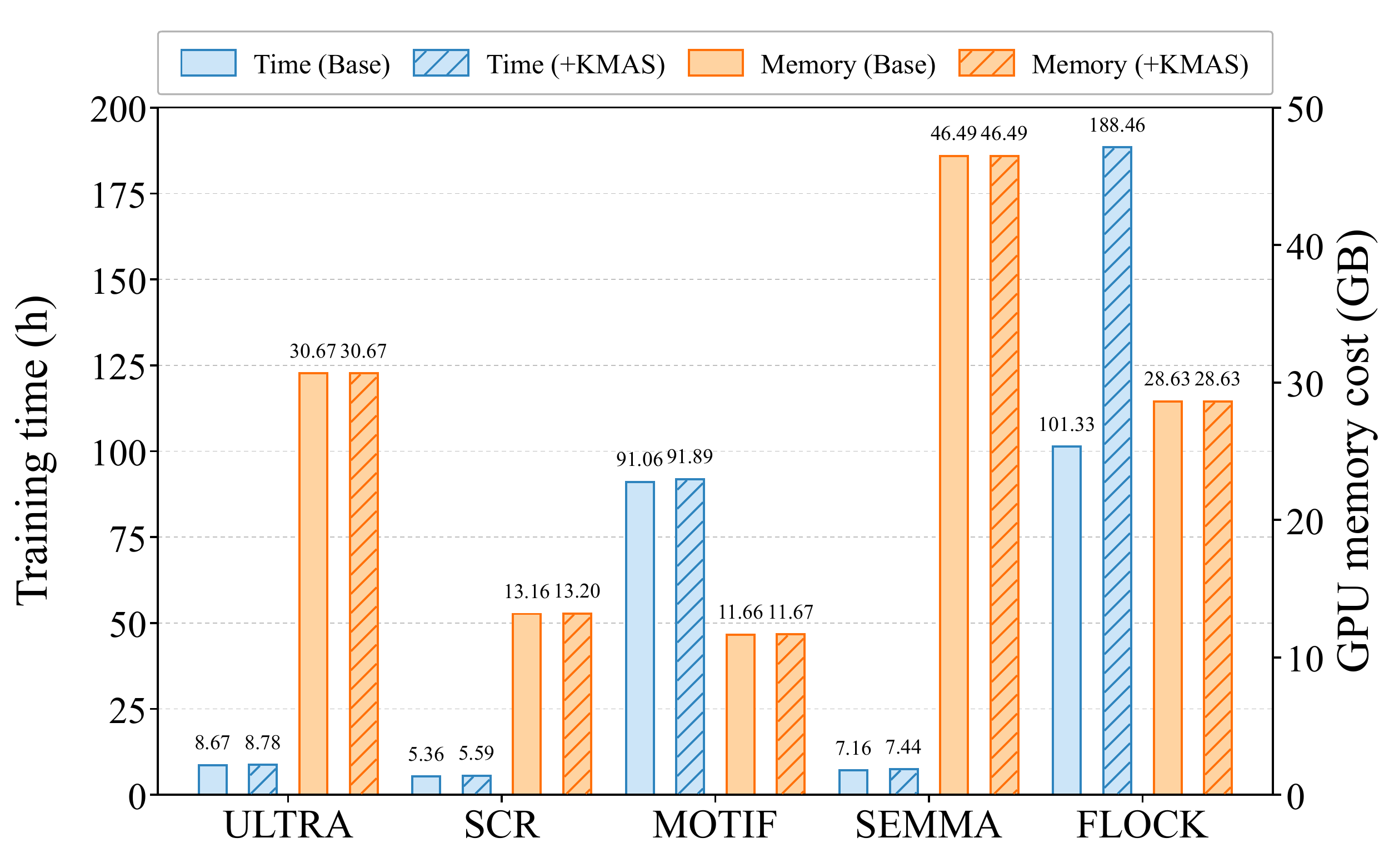}
    \vspace{-15pt}
    \caption{Cost: KGFMs vs. KGFMs+KMAS}
    \label{fig:cost_comparison}
    \vspace{-5pt}
\end{wrapfigure}
We test the training time for KMAS combined with different KGFMs on the same training data set (i.e., FB15k-237, WN18RR, and CoDEx-Medium). Since KMAS only performs in the training process, we do not show the inference time. The experimental results plotted in Figure~\ref{fig:cost_comparison} show that KMAS brings a very slight overhead to training time cost across GNN-based KGFMs (i.e., ULTRA,  MOTIF, and SEMMA), but it increases FLOCK's training time. For GNN-based KGFMs, KMAS only needs to obtain the relation embeddings generated by $\text{Encoder}_{\theta_r}$. 
As $\text{Encoder}_{\theta_e}$ accounts for the majority of the training cost, while $\text{Encoder}_{\theta_r}$ is computationally inexpensive, an extra run of $\text{Encoder}_{\theta_r}$ for the construction of negatives incurs only marginal overhead.
In contrast, the random walk-based KGFM, FLOCK, needs to perform the whole forward process to obtain the relation embeddings for the construction of negatives and use these constructed negatives to perform another forward process for training, which inevitably leads to nearly double the training cost. 

\subsection{Memory Cost Study} \label{sec:memeory_study}
 We test the memory cost for KMAS combined with different KGFMs on the same training data set (i.e., FB15k-237, WN18RR, and CoDEx-Medium) by tracking the peak memory usage via the \texttt{nvidia-smi} command. Experimental results (Figure~\ref{fig:cost_comparison}) show that KMAS does not incur significant additional memory overhead across different KGFMs (i.e., ULTRA, SCR, MOTIF, and SEMMA), demonstrating that KMAS is a memory-efficient method. These results are consistent with the memory complexity analysis in Section \ref{sec:comple}. Another point worth mentioning is that, despite FLOCK's entirely different architecture compared with those GNN-based KGFMs, KMAS still does not incur significant memory overhead on FLOCK.

\subsection{Ablation Study}
\label{sec:ablation}
\setlength{\columnsep}{5pt}

\begin{wraptable}[14]{r}{0.43\textwidth}
\vspace{-\intextsep}
\vspace{-9pt}
\centering

\caption{Ablation study results of KMAS.}
\label{tab:ablation}

\vspace{2pt}

\fontsize{8.5}{10}\selectfont
\setlength{\tabcolsep}{3pt}
\renewcommand{\arraystretch}{0.55}

\begin{adjustbox}{max width=\linewidth}
\begin{tabular}{ccc}
\toprule
\multirow{2}{*}{\bfseries\textit{Variant}}
& \multicolumn{2}{c}{\bfseries \textit {Total Avg (44 graphs)}} \\
\cmidrule(lr){2-3}
& \bfseries\textit{MRR}
& \bfseries\textit{Hits@10} \\
\midrule

$\text{ULTRA} + \text{KMAS}_{sub}$
& 0.364 & 0.525 \\

$\text{ULTRA} + \text{KMAS}_{hard}$
& 0.226 & 0.360 \\

$\text{ULTRA} + \text{KMAS}_{frozen}$
& 0.347 & 0.509 \\

$\text{ULTRA} + \text{KMAS}_{hold}$
& 0.362 & 0.525 \\

$\text{ULTRA} + \text{KMAS}_{increase}$
& 0.360 & 0.518 \\

$\text{ULTRA} + \text{KMAS}_{decrease}$
& 0.359 & 0.522 \\

\rowcolor{blue!10}
$\text{ULTRA} + \text{KMAS}$
& \bfseries 0.367 & \bfseries 0.532 \\

\midrule

$\text{SEMMA} + \text{KMAS}_{sub}$
& 0.373 & 0.527 \\

$\text{SEMMA} + \text{KMAS}_{hard}$
&  0.265 & 0.395 \\

$\text{SEMMA} + \text{KMAS}_{frozen}$
& 0.373 & 0.532 \\

$\text{SEMMA} + \text{KMAS}_{hold}$
& 0.371 & 0.527 \\

$\text{SEMMA} + \text{KMAS}_{increase}$
& 0.371 & 0.532 \\

$\text{SEMMA} + \text{KMAS}_{decrease}$
& 0.367 & 0.523 \\

\rowcolor{blue!10}
$\text{SEMMA} + \text{KMAS}$
& \bfseries 0.380 & \bfseries 0.539 \\

\bottomrule
\end{tabular}
\end{adjustbox}

\vspace{-4pt}
\end{wraptable}
We first construct 3 ablations: (1) $\text{KMAS}_{sub}$, replacing the construction of $\mathcal{T}^{hard}_{(h,r,t)}$ (resp. $\mathcal{H}^{hard}_{(h,r,t)}$) with random sampling; (2) $\text{KMAS}_{hard}$, using only hard negative samples after warmup, 
instead of combining hard and random negative samples; and (3) $\text{KMAS}_{frozen}$,freezing the relation embeddings obtained after warmup 
and using them for hard negatives construction throughout the remaining training process.
As shown in Table \ref{tab:ablation}, $\text{KMAS}_{sub}$ decreases the MRR/Hits@10 on ULTRA/SEMMA, demonstrating the effectiveness of negative sampling. $\text{KMAS}_{hard}$ drops further on ULTRA/SEMMA, indicating the importance of retaining random negatives in hybrid negative sampling. $\text{KMAS}_{frozen}$ also drops, showing that continuously updated relation embeddings are crucial for adapting negative sampling to the evolving state of KGFMs during training.

To evaluate the effect of the hard negative ratio adjustment strategy, we define $3$ variants: (1) $\text{KMAS}_{hold}$ (i.e., KMAS using the hard negative ratio $\alpha_{max}$ as $\alpha_{m}$ for the whole training process); (2) $\text{KMAS}_{increase}$ (i.e., KMAS increasing 
the hard negative ratio $\alpha_{m}$ from $0$ to $\alpha_{max}$ without decreasing); (3) $\text{KMAS}_{decrease}$ (i.e., KMAS decreasing $\alpha_{m}$ from $\alpha_{max}$ to $0$ without increasing). Experimental results (Table \ref{tab:ablation}) show that KMAS outperforms them, whether ULTRA or SEMMA is used as the base KGFM, demonstrating the superiority of our hard negative ratio adjustment strategy. 

\subsection{Analysis of Different Negative Sampling Methods}
\label{sec:two_comparison}
\setlength{\columnsep}{5pt}

\begin{wraptable}[12]{r}{0.43\textwidth}
\vspace{-\intextsep}
\vspace{-9pt}
\centering

\caption{Against different negative sampling methods.}
\label{tab:performance_comparison_ns}

\vspace{2pt}

\fontsize{8.5}{10}\selectfont
\setlength{\tabcolsep}{3pt}
\renewcommand{\arraystretch}{0.55}

\begin{adjustbox}{max width=\linewidth}
\begin{tabular}{ccc}
\toprule
\multirow{2}{*}{\bfseries\textit{Variant}}
& \multicolumn{2}{c}{\bfseries \textit {Total Avg (44 graphs)}} \\
\cmidrule(lr){2-3}
& \bfseries\textit{MRR}
& \bfseries\textit{Hits@10} \\
\midrule

$\text{ULTRA} + \text{SIMKGC}_{neg}$ (ACL'22)
& 0.277 & 0.408 \\

$\text{ULTRA} + \text{BIGRAM}_{neg}$ (TKDE'25)
& 0.282 & 0.421 \\

$\text{ULTRA} + \text{SATKGC}_{neg}$ (WWW'25)
& 0.318 & 0.461 \\

$\text{ULTRA} + \text{DANS}_{neg}$ (AAAI'26)
& 0.358 & 0.519 \\

$\text{ULTRA} + \text{ARNS}_{neg}$ (AAAI'26)
& 0.355 & 0.511 \\

\rowcolor{blue!10}
$\text{ULTRA} + \text{KMAS}$
& \bfseries 0.367 & \bfseries 0.532 \\

\midrule

$\text{SEMMA} + \text{SIMKGC}_{neg}$
& 0.360 & 0.522 \\

$\text{SEMMA} + \text{BIGRAM}_{neg}$
& 0.315 & 0.460 \\

$\text{SEMMA} + \text{SATKGC}_{neg}$
& 0.321 & 0.482 \\

$\text{SEMMA} + \text{DANS}_{neg}$
& 0.367 & 0.527 \\

$\text{SEMMA} + \text{ARNS}_{neg}$
& 0.365 & 0.514 \\

\rowcolor{blue!10}
$\text{SEMMA} + \text{KMAS}$
& \bfseries 0.380 & \bfseries 0.539 \\

\bottomrule
\end{tabular}
\end{adjustbox}

\vspace{-4pt}
\end{wraptable}

To further evaluate the effectiveness of KMAS, we provide the performance of two representative KGFMs (i.e., ULTRA and SEMMA) using five representative negative sampling methods (i.e., $\text{SIMKGC}_{neg}$, $\text{BIGRAM}_{neg}$, $\text{SATKGC}_{neg}$, $\text{DANS}_{neg}$ and $\text{ARNS}_{neg}$) over 44 data sets to compare with KMAS, as summarized in Table~\ref{tab:performance_comparison_ns}. $\text{SIMKGC}_{neg}$, $\text{BIGRAM}_{neg}$, $\text{SATKGC}_{neg}$, $\text{DANS}_{neg}$ and $\text{ARNS}_{neg}$ denote the negative sampling methods proposed by \citep{wang2022simkgc}, \citep{che2025hard}, \citep{ko2025subgraph}, \citep{li2026kgc} and \citep{ding2026arns}, respectively. Their poor performance may stem from their limited exploitation of cross-KG transferable invariances during negative sampling. KMAS leverages these invariances encoded in the KGFM’s continuously updated relation embeddings to construct hard negatives, thereby providing more informative supervision for the KGFM to further learn these invariances.

\subsection{Analysis of Statistical Significance }
\label{sec:significance}
\begin{wrapfigure}[13]{r}{0.35\textwidth}
\vspace{-\intextsep}
\vspace{-4pt}
\centering

\includegraphics[width=\linewidth]{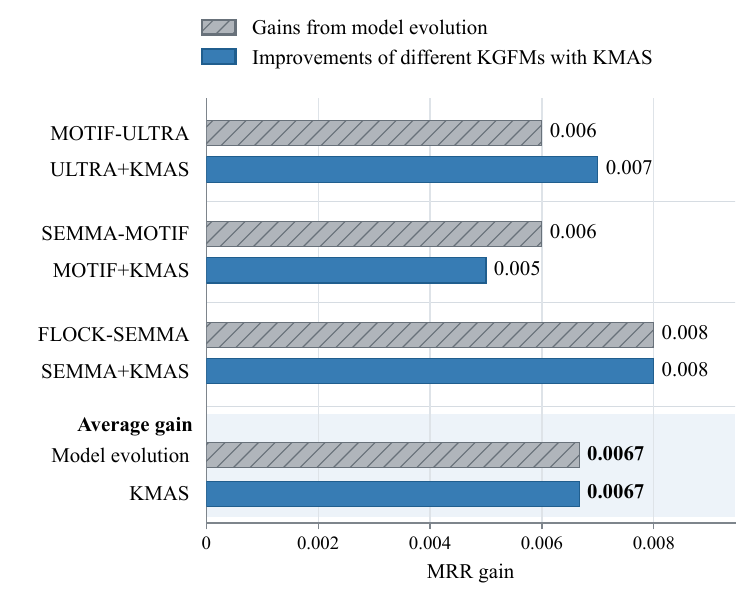}

\vspace{-6pt}

\caption{Average MRR gains from model evolution and KMAS.}
\label{fig:mrr_gains}

\vspace{-8pt}
\end{wrapfigure}
Following \citep{wiles2025revisiting, ghosh2025onebench, yu2025your}, we use the two-sided Wilcoxon signed-rank test \citep{wilcoxon1945individual} to assess the statistical significance of the average MRR improvements over 44 data sets. For some representative base KGFMs \text{ULTRA}, \text{SEMMA}, and \text{FLOCK}, this test yields $p$-values of 0.00264, 0.00375, and 0.00265, respectively. All the $p$-values fall below the conventional threshold of $0.05$, which means the improvements from KMAS are statistically significant and practically meaningful. 
In addition, as shown in Figure \ref{fig:mrr_gains}, the average MRR improvements from ULTRA to MOTIF, from MOTIF to SEMMA, and from SEMMA to FLOCK are $0.006$, $0.006$, and $0.008$, respectively, yielding an average model gain of $0.0067$.
In contrast, KMAS achieves average improvements of $0.007$, $0.005$, and $0.008$ over ULTRA, MOTIF, and SEMMA, respectively, with an average of $0.0067$. This indicates that the gains from KMAS are comparable to those from model evolution, verifying the practical value of KMAS.
\begin{figure}[htbp]
    \centering
    \includegraphics[width=0.48\linewidth]{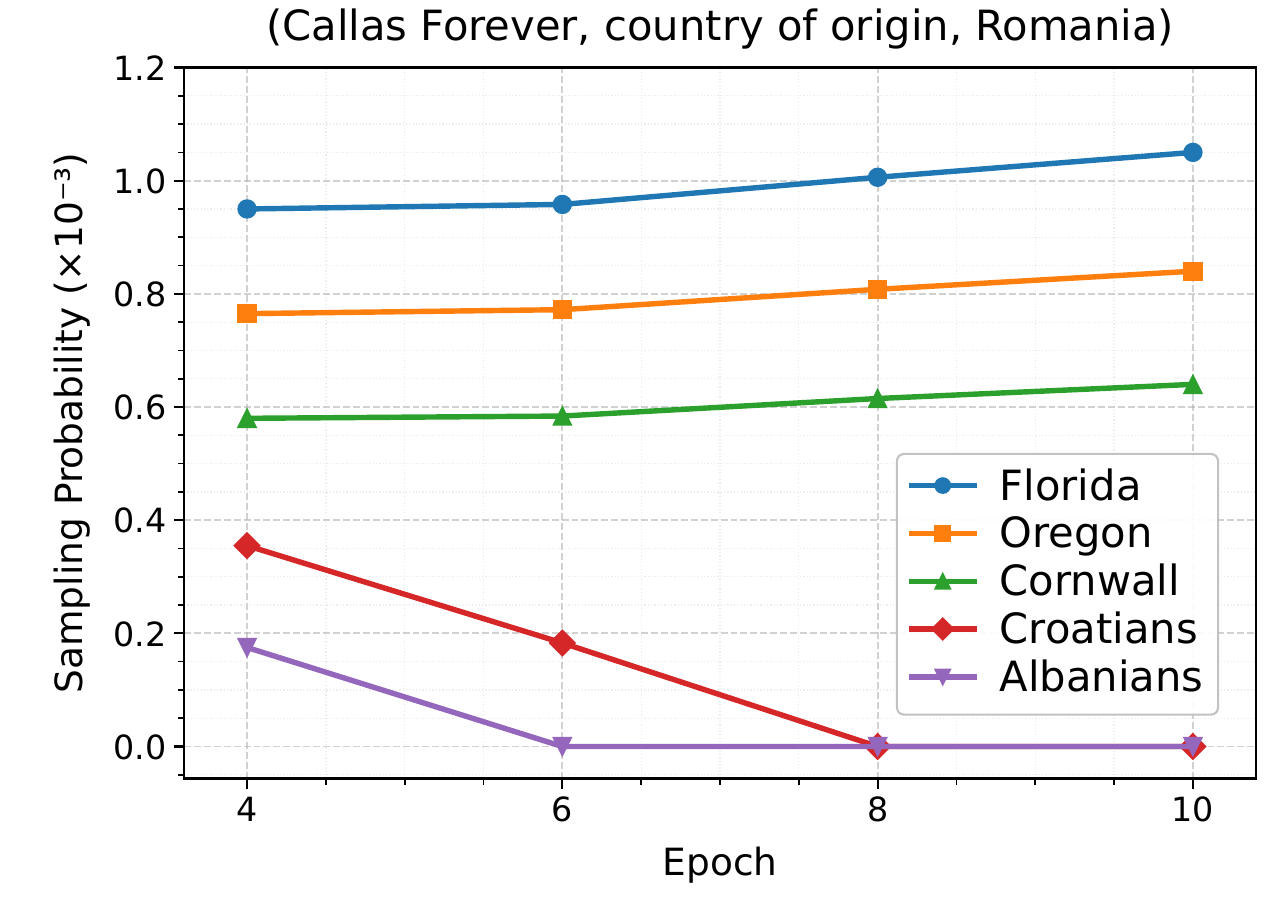}
    \hfill
    \includegraphics[width=0.48\linewidth]{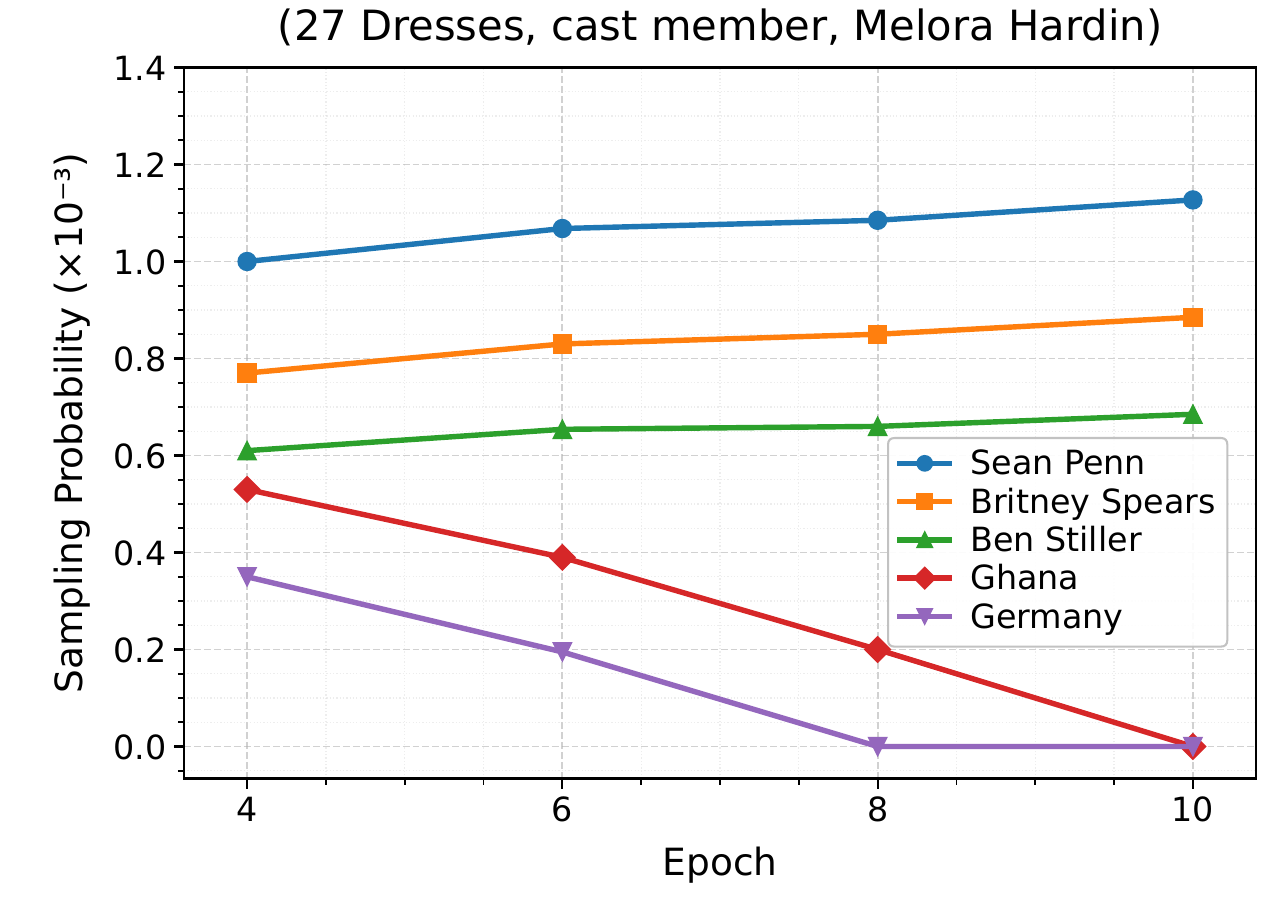}
    \caption{Sampling probability trends of negative
    candidates over training.}
    \label{fig:case_study}
\end{figure}
\subsection{Case Study}
\label{sec:case_study}
To verify KMAS's effectiveness in generating hard negatives, we conduct a case study on CoDEx-Medium with ULTRA as the base KGFM. For each of the 2 positive triples, we show 5 tail candidates used to construct negatives by replacing the true tail entity, along with their probability evolution at the start of the 4th, 6th, 8th, and 10th epochs. Taking (\textit{27 Dresses}, \textit{cast member}, \textit{Melora Hardin}) as the case (Figure \ref{fig:case_study}), KMAS dynamicly adjusts the sampling probabilities of candidate entities. Among 5 candidates, \textit{Sean Penn}, \textit{Britney Spears}, and \textit{Ben Stiller} are all person entities related to the entertainment domain, making them plausible yet still incorrect replacements for the true tail entity. Their sampling probabilities increase from the 4th to the 10th epoch, suggesting KMAS increasingly treats them as harder candidates and assigns them higher probabilities for negative construction. In contrast, \textit{Ghana} and \textit{Germany} are countries whose entity types are less compatible with the expected tail entity of the \textit{cast member} relation. Their probabilities consistently drop, suggesting KMAS suppresses easier or unsuitable candidates for the current relation. A similar trend appears in another example. This contrast shows that KMAS can dynamically adjust candidate probability distributions, constructing negatives more consistent with KGFM's evolving state.
\section{Related Work}
\noindent\textbf{Knowledge graph foundation models}
(KGFMs) have emerged as a paradigm for enabling zero-shot reasoning over unseen KGs by learning invariance of the relational structure across diverse KGs. ULTRA ~\cite{galkin2023ultra} pioneered KGFM through the construction of a relation graph $\mathcal{G}_R$ where edges are four fundamental interactions (i.e., \emph{h2h}, \emph{t2h}, \emph{h2t}, \emph{t2t}), enabling zero-shot inference on unseen entities and relations. MOTIF~\cite{huang2025how} provides a theoretical framework formalizing KGFMs based on graph motifs via 3 steps:
\textsc{Lift}, \textsc{Relation encoder}, and \textsc{Entity} \textsc{Encoder}. It reveals ULTRA's equivalence to binary 2-path motifs and finds that new motifs increase expressiveness only when not covered by core-onto homomorphisms.
SEMMA \cite{arun2025semmasemanticawareknowledge} extends KGFMs by combining structural and textual signals obtained from a text relation graph through an MLP to generate the final relation embeddings, empowering robust zero-shot link prediction across diverse KGs.
SCR \cite{wang2024towards} improves KGFMs by bridging the gap between structural patterns and node semantic features, while unifying diverse graph tasks into a single reasoning framework. 
FLOCK \cite{kim2025flock} performs diversified random walks from multiple starting locations over KGs, anonymizes them, uses a sequence model to encode these sequences to entity and relation embeddings, and aggregates them via a consensus protocol to obtain the final entity and relation embeddings. Notably, FLOCK relies on the random walk and the sequence model, while other KGFMs rely on GNN. Despite significant progress that has been made in KGFMs, the negative sample quality issue is ignored. All KGFMs above rely on corrupting either the head or tail entity of the positive sample to obtain negative samples where negative triples are independent of the invariance of the relation structure. Our work focuses on boosting KGFMs, which are based on relation graphs via an effective and simple negative sampling method.

\noindent\textbf{Negative sampling} methods serve as effective methods applied in the task of link prediction on KGs. 
\citet{wang2022simkgc} combines in-batch negatives, pre-batch negatives, and self-negatives, emphasizing hard negatives through the contrastive loss.
\citet{che2025hard} samples hard negative entities using a fixed distribution derived from relation and entity co-occurrence statistics through a bigram language model.
\citet{ko2025subgraph} constructs hard in-batch negatives from precomputed subgraphs and weights them by graph distance. It is noted that the above three methods do not adjust the hard negative ratio during training. 
\citet{li2026kgc} generates negatives through conditional diffusion guided by entity semantics and structure, with a curriculum that shifts sampling from easy to hard negatives.
\citet{ding2026arns} combines L-WD, relation-aware, and uniform random negatives, adjusting their mixture ratios based on MRR feedback.
 However, as verified in Section \ref{sec:two_comparison}, existing negative sampling methods cannot match the KGFM's evolving capability to distinguish between positive and negative samples, and directly adapting to KGFMs will degrade performance. Beyond simply integrating existing KG negative sampling techniques, KMAS has two novel aspects: (1) coupling of the KGFM relation encoder and negative generation process during each iteration for mutual reinforcement—a stronger encoder generates better relation embeddings containing cross‑KG transferable information, which guide cross‑KG transferable information‑aware negative sampling to construct negatives that strengthen the encoder and further boost the KGFM; (2) the increase-then-decrease hard‑negative ratio adjustment strategy, which is simple, effective, and plug‑and‑play—a clear departure from existing negative sampling techniques.

\noindent\textbf{Curriculum learning} \citep{bengio2009curriculum} is a training paradigm that progressively introduces examples from easy to hard. Its core idea is applied across various tasks, e.g., complex reasoning over KG \citep{xia2025improving,zhang2019iterative}, KG embedding \citep{guo2018knowledge}, and joint
open KG canonicalization and linking \citep{Shen2024CLUE}. Specifically, we propose a hard negative ratio adjustment strategy inspired by the trade-off policy, which combines the strategy of ``easier first" as curriculum learning and ``harder first" as hard example mining \citep{wang2021survey}.

\section{Conclusion}
To enhance existing KGFMs, we propose a simple yet effective
negative sampling method to generate hard negative samples whose ratio is dynamically adjusted to train KGFMs. KMAS is flexible to adapt to many KGFMs. Extensive experiments conducted on 44 data sets have demonstrated that KMAS can improve the performance of many SOTA KGFMs without requiring excessive additional memory consumption, and it incurs modest additional time cost on most KGFMs. In the future, we will evaluate the performance of KMAS on more KGFMs and data sets.

\section*{AI Use Statement}
In this work, we used generative AI tools to edit the manuscript to improve its language and readability. We did not use generative AI tools for other applicable tasks requiring disclosure.
\bibliography{iclr2027_conference}
\bibliographystyle{iclr2027_conference}

\appendix
\section{Algorithm}
\label{appendix:algorithm}
The process of KMAS is shown in Algorithm \ref{alg:RSWS-T}.
\begin{algorithm}[t]
\caption{KMAS}
\label{alg:RSWS-T}
\KwIn{KGs $\mathcal{G}_1, \mathcal{G}_2\ldots, \mathcal{G}_n$, relation graphs $\mathcal{G}_{R_1},\mathcal{G}_{R_2},\ldots,\mathcal{G}_{R_n}$, the entity encoder
$\text{Encoder}_{\theta_e}$, the relation encoder $\text{Encoder}_{\theta_{r}}$, the MLP $f_{\omega}$, the number of negatives $N$, threshold $ q $, temperatures $\tau$ and $\tau_{adv}$, total number of iterations $T_{\text{total}}$, warmup ratio $\beta_{\text{warm}}$, peak position $\gamma_{\text{peak}}$, maximum ratio of the hard negative samples $\alpha_{\max}$}

\KwOut{ Optimized parameters $\theta_{r}$, $\theta_e$, and $\omega$}

Initialize $\theta_{r}$, $\theta_{e}$, and $\omega$ randomly\\
\For{$m = 1$ to $T_{total}$}
{
    Randomly sample a KG $\mathcal{G}_i$ and a batch of positive triples $\mathcal{B}^+$ from $\mathcal{G}_i$\\
    
    Calculate the ratio of completed iterations $p_m$ 
    
    \For{each positive triple $(h, r, t) \in \mathcal{B}^+$ }
    {
        \If{$p_m < \beta_{\mathrm{warm}}$}{
            $\alpha_m \leftarrow 0$ 
        
            Draw $N_{rand}$  entities from $\mathcal{V}$ randomly

            Construct the negative sample set $\mathcal{B}^-_{(h, r, t)}$ for $(h, r, t)$ by replacing the head/tail entity with a random entity in $\mathcal{G}_i$
        }
        \Else{
        Obtain the relation embeddings $H$ \\
        Calculate the weight vector $W_r$ \\
        Generate the relation set $R_{topK}(r)$ based on $W_r$\\
        Calculate $W'$ \\
        Calculate the weights of entities ${x}^{tail}_{r, e}$ and $x^{head}_{r, e}$ based on $W'$ and $R_{topK}(r)$\\
        Generate entity probability distributions $\mathcal{P}^{tail}_r$ and $\mathcal{P}^{head}_{r}$ \\
        \If {$p_m < P_{\mathrm{peak}}$}{
            Update $\alpha_m$ 
        }
        \Else{
            Update $\alpha_m$ 
        }
        Draw $N_{rand}$ entities from $\mathcal{V}$ randomly\\
        Draw $N - N_{rand}$ entities according to $\mathcal{P}^{tail}_{r}$ and $\mathcal{P}^{head}_r$ \\
        Construct the negative sample set $\mathcal{B}^-_{(h, r, t)}$ for $(h, r, t)$ \\
        }
    }
    
    Calculate the loss based on $\mathcal{B}^+$ and its corresponding negative triple set $\mathcal{B}^- = \bigcup\limits_{(h ,r, t)  \in \mathcal{B}^+} \mathcal{B}^-_{(h, r, t)} $  \
    
    Update $\theta_{r}$, $\theta_{e}$, and $\omega$
}
\end{algorithm}

\section{KGFMs}
\label{appendix:kgfms}
We provide detailed descriptions of base models used in our experiments below:
\begin{itemize}
    \item ULTRA \cite{galkin2023ultra} constructs the relation graph, where nodes denote relations in the KG and edges denote four fundamental interactions between relations in the KG: 
     \emph{h2h}, \emph{t2t}, \emph{h2t}, and \emph{t2h}. An edge between relation $r_i$ and $r_j$ is established in $\mathcal{G}_\mathcal{R}$ if there exist triples $(h_i, r_i, t_i)$ and $(h_j, r_j, t_j) \in \mathcal{T}$ satisfying one of the following conditions: (1) \emph{h2h}: $h_i = h_j$; (2) \emph{t2t}: $t_i = t_j$; (3) \emph{h2t}: $h_i = t_j$; (4) \emph{t2h}: $t_i = h_j$. Specifically, the four fundamental interactions are learnable embeddings, which form part of $\theta_r$.   $\text{Encoder}_{\theta_{r}}$ works on the relation graph to generate relation embeddings. $\text{Encoder}_{\theta_{e}}$ uses these relation embeddings to generate entity embeddings for $f_{\omega}$ to predict entities.
     \item  MOTIF \cite{huang2025how} constructs the relation hypergraph, where hyperedges are matches of arbitrary graph motifs (i.e., 2-path and 3-path motifs) from the KG, and nodes are relations in the KG. $\text{Encoder}_{\theta_r}$ of MOTIF works on the hypergraph, and the arbitrary graph motifs are the learnable embeddings that are part of $\theta_r$ for $\text{Encoder}_{\theta_r}$. $\text{Encoder}_{\theta_e}$ and $f_{\omega}$ of MOTIF adopt the same mechanism as ULTRA.
     \item SCR \cite{wang2024towards} constructs the relation graph following ULTRA, where edges denote four types of relation level interactions: \emph{h2h}, \emph{t2t}, \emph{h2t}, and \emph{t2h}. $\text{Encoder}_{\theta_r}$ applies CMP to generate the query conditional relation embeddings $R_q$ and query independent embeddings $R_g$. At the entity level, SCMP incorporates the semantic neighbors of the query entity into the initialization and uses $R_q$ to obtain query conditional entity embeddings, while query independent CMP uses $R_g$ to obtain the semantic representation $H_g$. $H_g$ is then merged with the original CMP based embeddings through an MLP to obtain the final entity embeddings for $f_{\omega}$.
     \item SEMMA~\cite{arun2025semmasemanticawareknowledge} constructs a structural relation graph and a textual relation graph. For the textual relation graph, nodes denote relations, and weighted edges between them denote the cosine similarity of their textual embeddings. $\text{Encoder}_{\theta_r}$ of SEMMA works on the two relation graphs in parallel and uses an MLP to integrate their information and obtain the final relation embeddings. $\text{Encoder}_{\theta_e}$ and $f_{\omega}$ of SEMMA adopt the same mechanism as ULTRA.
     \item  FLOCK \cite{kim2025flock} is a recent KGFM based on the random walk technique, which is quite different from existing GNN-based KGFMs (i.e., ULTRA, MOTIF, and SEMMA). FLOCK performs diversified random walks from multiple starting locations over KGs to capture both local query context and global structural information. It then anonymizes the visited nodes and relations through a recording protocol to produce graph-agnostic sequences that preserve only structural roles. These sequences are encoded by a sequence model to derive entity and relation embeddings along with their confidence scores. Then these embeddings are aggregated via a consensus protocol with their confidence scores to update the final entity and relation embeddings for link prediction.
\end{itemize}

\section{Complexity Analysis of Time and Memory}
\label{appendix:tiam_mem}
The time complexity of KMAS is analyzed as follows. The time complexity of the relation similarity acquisition (Section \ref{sec:sm_acquire}) is $\mathcal{O}(|\mathcal{B}^+| \cdot |\mathcal{R}| \cdot d)$ (line 11 - 12 in Algorithm \ref{alg:RSWS-T}). The time complexity of the relation-aware entity distribution generation (Section \ref{sec:neg_gen}) is $\mathcal{O}(|\mathcal{B}^+| \cdot ( |\mathcal{R}| ( \log |\mathcal{R}| + \bar{d}_{rel} ) + \mathcal{|V|}) )$ (line 13 - 16 in Algorithm \ref{alg:RSWS-T}). The time complexity of the hybrid sampling strategy (Section \ref{sec:hybrid}) is $\mathcal{O}(\mathcal{|B^+|} \cdot N)$ (line 17 - 23 in Algorithm \ref{alg:RSWS-T}).
Consequently, the total time complexity for training is $\mathcal{O}( T_{total} \cdot \mathcal{|B^+|} \cdot (|\mathcal{R}|(d + \log |\mathcal{R}| + \bar{d}_{rel}) + |\mathcal{V}| + N))$.

The memory complexity of KMAS is analyzed as follows. The memory complexity of the relation similarity acquisition (Section \ref{sec:sm_acquire}) is $\mathcal{O}(|\mathcal{B^+}| \cdot |\mathcal{R}|)$ (line 11 - 12 in Algorithm \ref{alg:RSWS-T}). The memory complexity of the relation-aware entity distribution generation (Section \ref{sec:neg_gen}) is $\mathcal{O}(|\mathcal{B^+}| \cdot |\mathcal{V}|)$ (line 13 - 16 in Algorithm \ref{alg:RSWS-T}). The memory complexity of the hybrid negative sampling (Section \ref{sec:hybrid}) is $\mathcal{O}(\mathcal{|B^+|} \cdot N)$ (line 17 - 23 in Algorithm \ref{alg:RSWS-T}).
Consequently, the total memory complexity of our method is $\mathcal{O}( |\mathcal{B^+}| ( |\mathcal{R}| + |\mathcal{V}| +N ) ).$

\section{Data Sets}
\label{appendix:datasets}
We train KGFMs on 3 KGs and perform inference on 44 data sets. The statistics of 3 KGs for training in Table \ref{dataset:train_trans} and 9 data sets of the transductive setting for inference are shown in Table \ref{dataset:infre_trans}. The statistics of 16 data sets of the partially inductive setting (i.e., inductive (e)) for inference are shown in Table \ref{dataset:induc_e}. The statistics of 19 data sets of the fully inductive setting (i.e., inductive (e, r)) for inference are shown in Table \ref{dataset:indc_e_r}.

\begin{table}[ht]
\centering
\small
\renewcommand{\arraystretch}{1.2}
\caption{3 KGs for training. \textbf{E} denotes the number of entities. \textbf{R} denotes the number of relations. Train, Valid, and Test denote the number of triples for training, validation, and test in the corresponding data set. Task denotes the prediction task. h/t means the task of predicting both head entities and tail entities.
}
\label{dataset:train_trans}
\begin{tabular}{c c c c c c c c}
\toprule
\textbf{\textit{Data set}} & \textbf{\textit{E}} & \textbf{\textit{R}} & \textbf{\textit{Train}} & \textbf{\textit{Valid}} & \textbf{\textit{Test}} & \textbf{\textit{Task}} \\
\midrule
CoDExMedium                & 17050  & 51    & 185584  & 10310  & 10311   & h/t \\
FB15k237                    & 14541  & 237   & 272115  & 17535  & 20466  & h/t \\
WN18RR                        & 40943  & 11    & 86835   & 3034   & 3134   & h/t \\
\bottomrule
\end{tabular}
\label{tab:transductive}
\end{table}
\begin{table}[ht]
\centering
\small
\renewcommand{\arraystretch}{1.2}
\caption{9 data sets of the transductive setting for inference.
Task denotes the prediction task. h/t means the task of predicting both head entities and tail entities, tails means the task of predicting the tail entity.}
\label{dataset:infre_trans}
\begin{tabular}{c c c c c c c c}
\toprule
\textbf{\textit{Data set}} & \textbf{\textit{E}} & \textbf{\textit{R}} & \textbf{\textit{Train}} & \textbf{\textit{Valid}} & \textbf{\textit{Test}} & \textbf{\textit{Task}} \\
\midrule
CoDExSmall                  & 2034   & 42    & 32888   & 1827   & 1828   & h/t \\
CoDExLarge                  & 77951  & 69    & 551193  & 30622  & 30622  & h/t \\
NELL995                          & 74536  & 200   & 149678  & 543   & 2818   & h/t \\
WDsinger                            & 10282  & 135   & 16142   & 2163   & 2203   & h/t \\
NELL23k                            & 22925  & 200   & 25445   & 4961   & 4952   & h/t \\
FB15k237\_10                        & 11512  & 237   & 27211   & 15624  & 18150  & tails \\
FB15k237\_20                       & 13166  & 237   & 54423   & 16963  & 19776  & tails \\
FB15k237\_50                        & 14149  & 237   & 136057  & 17449  & 20324  & tails \\
Hetionet                   & 45158  & 24    & 2025177 & 112510 & 112510 & h/t \\
\bottomrule
\end{tabular}
\label{tab:transductive}
\end{table}
\begin{table*}[ht]
\centering
\small
\renewcommand{\arraystretch}{1.2}
\caption{16 data sets of the partially inductive setting. T denotes the number of triples in their corresponding graph. Valid and Test denote the triples that need to be predicted
in the validation and test data sets in their corresponding graph. We predict both head entities and tail
entities for these data sets.}
\label{dataset:induc_e}
\begin{adjustbox}{max width=\linewidth}
\begin{tabular}{c  c  cc  ccc  ccc}
\toprule
\multirow{2}{*}{\bf{\textit{Data set}}} & 
\multirow{2}{*}{\bf{\textit{R}}}  
& \multicolumn{2}{c}{\bf{\textit{Train Graph}}} 
& \multicolumn{3}{c}{\bf{\textit{Validation Graph}}} 
& \multicolumn{3}{c}{\bf{\textit{Test Graph}}} \\
\cmidrule(lr){3-4} \cmidrule(lr){5-7} \cmidrule(lr){8-10}
 & & \bf{\textit{E}} & \bf{\textit{T}} 
   & \bf{\textit{E}} & \bf{\textit{T}} & \bf{\textit{Valid}} 
   & \bf{\textit{E}} & \bf{\textit{T}} & \bf{\textit{Test}} \\
\midrule
WN18RRInductive:v1      & 9   & 2746  & 5410   & 2746  & 5410   & 630   & 922   & 1618   & 373 \\
WN18RRInductive:v2      & 10  & 6954  & 15262  & 6954  & 15262  & 1838  & 2757  & 4011   & 852 \\
WN18RRInductive:v3      & 11  & 12078 & 25901  & 12078 & 25901  & 3097  & 5084  & 6327   & 1143 \\
WN18RRInductive:v4      & 9   & 3861  & 7940   & 3861  & 7940   & 934   & 7084  & 12334  & 2823 \\
FB15k237Inductive:v1      & 180 & 1594  & 4245   & 1594  & 4245   & 489   & 1093  & 1993   & 411 \\
FB15k237Inductive:v2      & 200 & 2608  & 9739   & 2608  & 9739   & 1166  & 1660  & 4145   & 947 \\
FB15k237Inductive:v3      & 215 & 3668  & 17986  & 3668  & 17986  & 2194  & 2501  & 7406   & 1731 \\
FB15k237Inductive:v4      & 219 & 4707  & 27203  & 4707  & 27203  & 3352  & 3051  & 11714  & 2840 \\
NELLInductive:v2    & 88  & 2564  & 8219   & 2564  & 8219   & 922   & 2086  & 4586   & 935 \\
NELLInductive:v3    & 142 & 4647  & 16393  & 4647  & 16393  & 1851  & 3566  & 8048   & 1620 \\
NELLInductive:v4    & 76  & 2092  & 7546   & 2092  & 7546   & 876   & 2795  & 7073   & 1447 \\
ILPC2022:small  & 48  & 10230 & 78616  & 6653  & 20960  & 2908  & 6653  & 20960  & 2902 \\
ILPC2022:large  & 65  & 46626 & 202446 & 29246 & 77044  & 10179 & 29246 & 77044  & 10184 \\
HM 3k     & 11  & 32118 & 71097  & 32250 & 71097  & 1201  & 19218 & 38285  & 1349 \\
HM 5k     & 11  & 28601 & 57601  & 28744 & 57601  & 900   & 23792 & 48425  & 2124 \\
HM:indigo  & 229 & 12721 & 121601 & 12797 & 121601 & 14121 & 14775 & 250195 & 14904 \\
\bottomrule
\end{tabular}
\end{adjustbox}
\label{tab:inductive_entity_datasets}
\end{table*}
\begin{table*}[ht]
\centering
\small
\renewcommand{\arraystretch}{1.2}
\caption{19 data sets of the fully inductive setting. 
We predict both head entities and tail entities for these data sets.}
\label{dataset:indc_e_r}
\begin{adjustbox}{max width=\linewidth}
\begin{tabular}{c ccc  cccc  cccc}
\toprule
\multirow{2}{*}{\bf{\textit{Data set}}} 
& \multicolumn{3}{c}{\bf{\textit{Train Graph}}} 
& \multicolumn{4}{c}{\bf{\textit{Validation Graph}}} 
& \multicolumn{4}{c}{\bf{\textit{Test Graph}}} \\
\cmidrule(lr){2-4} \cmidrule(lr){5-8} \cmidrule(lr){9-12}
& \bf{\textit{E}} & \bf{\textit{R}} & \bf{\textit{T}} 
& \bf{\textit{E}} & \bf{\textit{R}} & \bf{\textit{T}} & \bf{\textit{Valid}} 
& \bf{\textit{E}} & \bf{\textit{R}} & \bf{\textit{T}} & \bf{\textit{Test}} \\
\midrule
FBIngram:100          & 4659  & 134 & 62809  & 2624  & 77  & 6987  & 2329  & 2624  & 77  & 6987  & 2329 \\
FBIngram:75            & 4659  & 134 & 62809  & 2792  & 186 & 9316  & 3106  & 2792  & 186 & 9316  & 3106 \\
FBIngram:50          & 5190  & 153 & 85375  & 4445  & 205 & 11636 & 3879  & 4445  & 205 & 11636 & 3879 \\
FBIngram:25           & 5190  & 163 & 91571  & 4097  & 216 & 17147 & 5716  & 4097  & 216 & 17147 & 5716 \\
WKIngram:100          & 9784  & 67  & 49875  & 12136 & 37  & 13487 & 4496  & 12136 & 37  & 13487 & 4496 \\
WKIngram:75           & 6853  & 52  & 28741  & 2722  & 65  & 3430  & 1143  & 2722  & 65  & 3430  & 1144 \\
WKIngram:50           & 12022 & 72  & 82481  & 9328  & 93  & 9672  & 3224  & 9328  & 93  & 9672  & 3225 \\
WKIngram:25            & 12659 & 47  & 41873  & 3228  & 74  & 3391  & 1130  & 3228  & 74  & 3391  & 1131 \\
NLIngram:75            & 2607  & 96  & 11058  & 1578  & 116 & 1818  & 606   & 1578  & 116 & 1818  & 607 \\
NLIngram:50           & 4396  & 106 & 17578  & 2335  & 119 & 2576  & 859   & 2335  & 119 & 2576  & 859 \\
NLIngram:25            & 4396  & 106 & 17578  & 2146  & 120 & 2230  & 743   & 2146  & 120 & 2230  & 744 \\
MT1 tax    & 10000 & 10  & 17178  & 10000 & 10  & 17178 & 1908  & 10000 & 9   & 16526 & 1834 \\
MT2 org     & 10000 & 10  & 23233  & 10000 & 10  & 23233 & 2581  & 10000 & 11  & 21976 & 2441 \\
MT2 sci    & 10000 & 16  & 16471  & 10000 & 16  & 16471 & 1830  & 10000 & 16  & 14852 & 1650 \\
MT3 art    & 10000 & 45  & 27262  & 10000 & 45  & 27262 & 3026  & 10000 & 45  & 28023 & 3113 \\
MT3 infra  & 10000 & 24  & 21990  & 10000 & 24  & 21990 & 2443  & 10000 & 27  & 21646 & 2405 \\
MT4 sci    & 10000 & 42  & 12576  & 10000 & 42  & 12576 & 1397  & 10000 & 42  & 12516 & 1388 \\
Metafam         & 1316  & 28  & 13821  & 1316  & 28  & 13821 & 590   & 656   & 28  & 7257  & 184 \\
FBNELL          & 4636  & 100 & 10275  & 4636  & 100 & 10275 & 1055  & 4752  & 183 & 10685 & 597 \\
\bottomrule
\end{tabular}
\end{adjustbox}
\end{table*}

\section{Detailed Results}
\label{appendix:full_res}
We show detailed results of five KGFMs (i.e., ULTRA, SCR, MOTIF, SEMMA and FLOCK) and these KGFMs enhanced by KMAS over 44 data sets in Table \ref{res:ultra_transductive}, Table \ref{res:scr_transductive}, Table \ref{res:motif_transductive}, Table \ref{res:semma_transductive} and \ref{res:flock_transductive} corresponding to the results in Table \ref{tab:my_comparison}.

\begin{table}[!p]
\centering
\caption{Detail results of the KGFM ULTRA and ULTRA using KMAS as the negative sampling method.}
\label{res:ultra_transductive}
\begin{tabular}{ccccc}
\toprule
\multirow{2}{*}{\bf{\textit{Data set}}} & \multicolumn{2}{c}{ULTRA} & \multicolumn{2}{c}{ULTRA + KMAS} \\ \cmidrule(l){2-3} \cmidrule(l){4-5} 
 & \bf{\textit{MRR}} & \bf{\textit{Hits@10}} & \bf{\textit{MRR}} & \bf{\textit{Hits@10}} \\
\midrule
\multicolumn{5}{c}{\textit{Transductive}} \\
\midrule
CoDExSmall & 0.473 & 0.670 & 0.476 & 0.670 \\
CoDExLarge & 0.340 & 0.474 & 0.339 & 0.471 \\
NELL995 & 0.472 & 0.594 & 0.476 & 0.605 \\
WDsinger & 0.351 & 0.480 & 0.387 & 0.504 \\
NELL23k & 0.235 & 0.402 & 0.235 & 0.408 \\
FB15k237\_10 & 0.235 & 0.380 & 0.240 & 0.391 \\
FB15k237\_20 & 0.265 & 0.423 & 0.268 & 0.431 \\
FB15k237\_50 & 0.323 & 0.519 & 0.324 & 0.523 \\
Hetionet & 0.284 & 0.406 & 0.284 & 0.406 \\
\midrule
\multicolumn{5}{c}{\textit{Inductive (e) }} \\
\midrule
WN18RRInductive:v1 & 0.658 & 0.776 & 0.652 & 0.794 \\
WN18RRInductive:v2 & 0.658 & 0.772 & 0.657 & 0.771 \\
WN18RRInductive:v3 & 0.373 & 0.496 & 0.383 & 0.510 \\
WN18RRInductive:v4 & 0.605 & 0.714 & 0.605 & 0.716 \\
FB15k237Inductive:v1 & 0.484 & 0.645 & 0.495 & 0.647 \\
FB15k237Inductive:v2 & 0.486 & 0.684 & 0.491 & 0.693 \\
FB15k237Inductive:v3 & 0.483 & 0.650 & 0.485 & 0.646 \\
FB15k237Inductive:v4 & 0.483 & 0.672 & 0.482 & 0.674 \\
NELLInductive:v2 & 0.499 & 0.696 & 0.536 & 0.720 \\
NELLInductive:v3 & 0.521 & 0.706 & 0.530 & 0.713 \\
NELLInductive:v4 & 0.477 & 0.711 & 0.495 & 0.729 \\
ILPC2022:small & 0.292 & 0.444 & 0.293 & 0.452 \\
ILPC2022:large & 0.283 & 0.420 & 0.295 & 0.424 \\
HM:3k & 0.061 & 0.107 & 0.054 & 0.109 \\
HM:5k & 0.054 & 0.102 & 0.050 & 0.099 \\
HM:indigo & 0.440 & 0.649 & 0.436 & 0.646 \\
\midrule
\multicolumn{5}{c}{\textit{Inductive (e, r)}} \\
\midrule
FBIngram:100 & 0.439 & 0.627 & 0.439 & 0.632 \\
FBIngram:75 & 0.394 & 0.595 & 0.400 & 0.599 \\
FBIngram:50 & 0.329 & 0.532 & 0.333 & 0.541 \\
FBIngram:25 & 0.387 & 0.634 & 0.392 & 0.639 \\
WKIngram:100 & 0.178 & 0.291 & 0.184 & 0.302 \\
WKIngram:75 & 0.373 & 0.520 & 0.379 & 0.536 \\
WKIngram:50 & 0.143 & 0.295 & 0.151 & 0.308 \\
WKIngram:25 & 0.286 & 0.498 & 0.297 & 0.512 \\
NLIngram:75 & 0.347 & 0.513 & 0.325 & 0.502 \\
NLIngram:50 & 0.382 & 0.559 & 0.376 & 0.532 \\
NLIngram:25 & 0.381 & 0.556 & 0.371 & 0.558 \\
WikiTopicsMT1:tax & 0.223 & 0.308 & 0.233 & 0.320 \\
WikiTopicsMT2:org & 0.088 & 0.153 & 0.090 & 0.153 \\
WikiTopicsMT2:sci & 0.236 & 0.390 & 0.269 & 0.428 \\
WikiTopicsMT3:art & 0.244 & 0.409 & 0.261 & 0.422 \\
WikiTopicsMT3:infra & 0.640 & 0.779 & 0.636 & 0.780 \\
WikiTopicsMT4:sci & 0.298 & 0.455 & 0.299 & 0.460 \\
Metafam & 0.177 & 0.726 & 0.267 & 0.788 \\
FBNELL & 0.479 & 0.646 & 0.482 & 0.646 \\
\bottomrule
\end{tabular}
\end{table}
\begin{table}[!p]
\centering
\caption{Detailed results of SCR and SCR using KMAS as the negative sampling method.}
\label{res:scr_transductive}
\begin{tabular}{ccccc}
\toprule
\multirow{2}{*}{\bf{\textit{Data set}}} & \multicolumn{2}{c}{SCR} & \multicolumn{2}{c}{SCR + KMAS} \\ \cmidrule(l){2-3} \cmidrule(l){4-5} 
 & \bf{\textit{MRR}} & \bf{\textit{Hits@10}} & \bf{\textit{MRR}} & \bf{\textit{Hits@10}} \\
\midrule
\multicolumn{5}{c}{\textit{Transductive}} \\
\midrule
CoDExSmall & 0.400 & 0.601 & 0.415 & 0.606 \\
CoDExLarge & 0.320 & 0.436 & 0.320 & 0.441 \\
NELL995 & 0.451 & 0.584 & 0.491 & 0.614 \\
WDsinger & 0.354 & 0.462 & 0.371 & 0.480 \\
NELL23k & 0.231 & 0.399 & 0.225 & 0.397 \\
FB15k237\_10 & 0.235 & 0.374 & 0.228 & 0.368 \\
FB15k237\_20 & 0.261 & 0.421 & 0.263 & 0.420 \\
FB15k237\_50 & 0.317 & 0.507 & 0.315 & 0.512 \\
Hetionet & 0.321 & 0.440 & 0.267 & 0.403 \\
\midrule
\multicolumn{5}{c}{\textit{Inductive (e)}} \\
\midrule
WN18RRInductive:v1 & 0.559 & 0.732 & 0.595 & 0.760 \\
WN18RRInductive:v2 & 0.564 & 0.741 & 0.606 & 0.762 \\
WN18RRInductive:v3 & 0.337 & 0.478 & 0.345 & 0.490 \\
WN18RRInductive:v4 & 0.479 & 0.650 & 0.540 & 0.679 \\
FB15k237Inductive:v1 & 0.475 & 0.646 & 0.465 & 0.652 \\
FB15k237Inductive:v2 & 0.484 & 0.688 & 0.490 & 0.687 \\
FB15k237Inductive:v3 & 0.479 & 0.648 & 0.483 & 0.650 \\
FB15k237Inductive:v4 & 0.474 & 0.667 & 0.483 & 0.670 \\
NELLInductive:v2 & 0.536 & 0.738 & 0.549 & 0.754 \\
NELLInductive:v3 & 0.535 & 0.708 & 0.542 & 0.731 \\
NELLInductive:v4 & 0.493 & 0.726 & 0.495 & 0.733 \\
ILPC2022:small & 0.276 & 0.414 & 0.281 & 0.423 \\
ILPC2022:large & 0.288 & 0.408 & 0.293 & 0.416 \\
HM:3k & 0.058 & 0.091 & 0.037 & 0.070 \\
HM:5k & 0.047 & 0.073 & 0.037 & 0.065 \\
HM:indigo & 0.422 & 0.635 & 0.414 & 0.625 \\
\midrule
\multicolumn{5}{c}{\textit{Inductive (e, r)}} \\
\midrule
FBIngram:100 & 0.426 & 0.625 & 0.427 & 0.629 \\
FBIngram:75 & 0.384 & 0.586 & 0.390 & 0.591 \\
FBIngram:50 & 0.322 & 0.520 & 0.331 & 0.535 \\
FBIngram:25 & 0.375 & 0.622 & 0.382 & 0.632 \\
WKIngram:100 & 0.176 & 0.283 & 0.179 & 0.291 \\
WKIngram:75 & 0.371 & 0.523 & 0.366 & 0.510 \\
WKIngram:50 & 0.151 & 0.294 & 0.145 & 0.275 \\
WKIngram:25 & 0.283 & 0.476 & 0.271 & 0.455 \\
NLIngram:75 & 0.344 & 0.521 & 0.336 & 0.535 \\
NLIngram:50 & 0.380 & 0.559 & 0.376 & 0.568 \\
NLIngram:25 & 0.362 & 0.548 & 0.356 & 0.540 \\
WikiTopicsMT1:tax & 0.220 & 0.301 & 0.225 & 0.297 \\
WikiTopicsMT2:org & 0.066 & 0.114 & 0.060 & 0.108 \\
WikiTopicsMT2:sci & 0.237 & 0.342 & 0.251 & 0.358 \\
WikiTopicsMT3:art & 0.247 & 0.383 & 0.243 & 0.393 \\
WikiTopicsMT3:infra & 0.600 & 0.738 & 0.597 & 0.735 \\
WikiTopicsMT4:sci & 0.272 & 0.427 & 0.272 & 0.444 \\
Metafam & 0.251 & 0.649 & 0.239 & 0.576 \\
FBNELL & 0.467 & 0.633 & 0.475 & 0.667 \\
\bottomrule
\end{tabular}
\end{table}

\begin{table}[!p]
\centering
\caption{Detail results of the KGFM MOTIF and MOTIF using KMAS as the negative sampling method.} 
\label{res:motif_transductive}
\begin{tabular}{ccccc}
\toprule
\multirow{2}{*}{\bf{\textit{Data set}}} & \multicolumn{2}{c}{MOTIF} & \multicolumn{2}{c}{MOTIF + KMAS} \\ \cmidrule(l){2-3} \cmidrule(l){4-5} 
 & \bf{\textit{MRR}} & \bf{\textit{Hits@10}} & \bf{\textit{MRR}} & \bf{\textit{Hits@10}} \\
\midrule
\multicolumn{5}{c}{\textit{Transductive}} \\
\midrule
CoDExSmall & 0.472 & 0.668 & 0.470 & 0.667 \\
CoDExLarge & 0.331 & 0.462 & 0.340 & 0.471 \\
NELL995 & 0.440 & 0.596 & 0.468 & 0.616 \\
WDsinger & 0.372 & 0.485 & 0.379 & 0.498 \\
NELL23k & 0.220 & 0.374 & 0.223 & 0.389 \\
FB15k237\_10 & 0.244 & 0.390 & 0.244 & 0.390 \\
FB15k237\_20 & 0.263 & 0.421 & 0.266 & 0.424 \\
FB15k237\_50 & 0.313 & 0.510 & 0.318 & 0.514 \\
Hetionet & 0.264 & 0.386 & 0.284 & 0.406 \\
\midrule
\multicolumn{5}{c}{\textit{ Inductive (e)}} \\
\midrule
WN18RRInductive:v1 & 0.663 & 0.777 & 0.686 & 0.787 \\
WN18RRInductive:v2 & 0.663 & 0.769 & 0.658 & 0.773 \\
WN18RRInductive:v3 & 0.430 & 0.565 & 0.425 & 0.547 \\
WN18RRInductive:v4 & 0.629 & 0.705 & 0.630 & 0.709 \\
FB15k237Inductive:v1 & 0.485 & 0.669 & 0.484 & 0.681 \\
FB15k237Inductive:v2 & 0.494 & 0.719 & 0.494 & 0.715 \\
FB15k237Inductive:v3 & 0.486 & 0.664 & 0.471 & 0.663 \\
FB15k237Inductive:v4 & 0.480 & 0.673 & 0.478 & 0.668 \\
NELLInductive:v2 & 0.527 & 0.724 & 0.532 & 0.742 \\
NELLInductive:v3 & 0.504 & 0.678 & 0.521 & 0.696 \\
NELLInductive:v4 & 0.475 & 0.685 & 0.471 & 0.731 \\
ILPC2022:small & 0.286 & 0.439 & 0.292 & 0.445 \\
ILPC2022:large & 0.274 & 0.416 & 0.269 & 0.413 \\
HM:3k & 0.059 & 0.100 & 0.100 & 0.107 \\
HM:5k & 0.050 & 0.087 & 0.056 & 0.099 \\
HM:indigo & 0.428 & 0.638 & 0.431 & 0.640 \\
\midrule
\multicolumn{5}{c}{\textit{Inductive (e, r)}} \\
\midrule
FBIngram:100 & 0.437 & 0.629 & 0.428 & 0.625 \\
FBIngram:75 & 0.394 & 0.601 & 0.389 & 0.601 \\
FBIngram:50 & 0.332 & 0.528 & 0.336 & 0.542 \\
FBIngram:25 & 0.380 & 0.626 & 0.388 & 0.637 \\
WKIngram:100 & 0.174 & 0.294 & 0.164 & 0.288 \\
WKIngram:75 & 0.366 & 0.531 & 0.368 & 0.518 \\
WKIngram:50 & 0.165 & 0.318 & 0.165 & 0.311 \\
WKIngram:25 & 0.321 & 0.514 & 0.309 & 0.506 \\
NLIngram:75 & 0.324 & 0.488 & 0.327 & 0.498 \\
NLIngram:50 & 0.380 & 0.527 & 0.374 & 0.538 \\
NLIngram:25 & 0.326 & 0.454 & 0.336 & 0.519 \\
WikiTopicsMT1:tax & 0.244 & 0.342 & 0.258 & 0.408 \\
WikiTopicsMT2:org & 0.094 & 0.153 & 0.090 & 0.153 \\
WikiTopicsMT2:sci & 0.275 & 0.422 & 0.266 & 0.428 \\
WikiTopicsMT3:art & 0.259 & 0.409 & 0.263 & 0.412 \\
WikiTopicsMT3:infra & 0.620 & 0.764 & 0.638 & 0.772 \\
WikiTopicsMT4:sci & 0.287 & 0.462 & 0.276 & 0.429 \\
Metafam & 0.417 & 0.764 & 0.488 & 0.807 \\
FBNELL & 0.463 & 0.633 & 0.468 & 0.651 \\
\bottomrule
\end{tabular}
\end{table}
\begin{table}[!p]
\centering
\caption{Detail results of the KGFM SEMMA and SEMMA using KMAS as the negative sampling method.}
\label{res:semma_transductive}
\begin{tabular}{ccccc}
\toprule
\multirow{2}{*}{\bf{\textit{Data set}}} & \multicolumn{2}{c}{SEMMA} & \multicolumn{2}{c}{SEMMA + KMAS} \\ \cmidrule(l){2-3} \cmidrule(l){4-5} 
 & \bf{\textit{MRR}} & \bf{\textit{Hits@10}} & \bf{\textit{MRR}} & \bf{\textit{Hits@10}} \\
\midrule
\multicolumn{5}{c}{\textit{Transductive}} \\
\midrule
CoDExSmall & 0.489 & 0.676 & 0.484 & 0.679 \\
CoDExLarge & 0.346 & 0.479 & 0.337 & 0.473 \\
NELL995 & 0.416 & 0.546 & 0.468 & 0.604 \\
WDsinger & 0.390 & 0.503 & 0.395 & 0.502 \\
NELL23k & 0.241 & 0.413 & 0.238 & 0.412 \\
FB15k237\_10 & 0.233 & 0.385 & 0.241 & 0.385 \\
FB15k237\_20 & 0.262 & 0.430 & 0.266 & 0.427 \\
FB15k237\_50 & 0.326 & 0.524 & 0.329 & 0.525 \\
Hetionet & 0.237 & 0.356 & 0.264 & 0.375 \\
\midrule
\multicolumn{5}{c}{\textit{ Inductive (e)}} \\
\midrule
WN18RRInductive:v1 & 0.709 & 0.806 & 0.702 & 0.808 \\
WN18RRInductive:v2 & 0.695 & 0.795 & 0.702 & 0.800 \\
WN18RRInductive:v3 & 0.445 & 0.591 & 0.445 & 0.592 \\
WN18RRInductive:v4 & 0.658 & 0.739 & 0.654 & 0.730 \\
FB15k237Inductive:v1 & 0.495 & 0.647 & 0.504 & 0.656 \\
FB15k237Inductive:v2 & 0.513 & 0.696 & 0.508 & 0.695 \\
FB15k237Inductive:v3 & 0.496 & 0.654 & 0.504 & 0.658 \\
FB15k237Inductive:v4 & 0.498 & 0.674 & 0.496 & 0.681 \\
NELLInductive:v2 & 0.526 & 0.711 & 0.531 & 0.728 \\
NELLInductive:v3 & 0.497 & 0.691 & 0.525 & 0.715 \\
NELLInductive:v4 & 0.480 & 0.726 & 0.492 & 0.727 \\
ILPC2022:small & 0.303 & 0.455 & 0.301 & 0.452 \\
ILPC2022:large & 0.303 & 0.426 & 0.304 & 0.426 \\
HM:3k & 0.053 & 0.084 & 0.057 & 0.094 \\
HM:5k & 0.051 & 0.096 & 0.051 & 0.086 \\
HM:indigo & 0.430 & 0.643 & 0.433 & 0.644 \\
\midrule
\multicolumn{5}{c}{\textit{Inductive (e, r)}} \\
\midrule
FBIngram:100 & 0.454 & 0.643 & 0.452 & 0.639 \\
FBIngram:75 & 0.403 & 0.598 & 0.409 & 0.607 \\
FBIngram:50 & 0.342 & 0.548 & 0.341 & 0.546 \\
FBIngram:25 & 0.399 & 0.644 & 0.400 & 0.649 \\
WKIngram:100 & 0.185 & 0.295 & 0.187 & 0.303 \\
WKIngram:75 & 0.382 & 0.526 & 0.381 & 0.515 \\
WKIngram:50 & 0.173 & 0.319 & 0.175 & 0.322 \\
WKIngram:25 & 0.312 & 0.481 & 0.308 & 0.489 \\
NLIngram:75 & 0.349 & 0.523 & 0.351 & 0.528 \\
NLIngram:50 & 0.394 & 0.573 & 0.386 & 0.550 \\
NLIngram:25 & 0.382 & 0.532 & 0.388 & 0.573 \\
WikiTopicsMT1:tax & 0.233 & 0.310 & 0.252 & 0.398 \\
WikiTopicsMT2:org & 0.095 & 0.158 & 0.095 & 0.159 \\
WikiTopicsMT2:sci & 0.237 & 0.340 & 0.258 & 0.398 \\
WikiTopicsMT3:art & 0.270 & 0.416 & 0.277 & 0.423 \\
WikiTopicsMT3:infra & 0.634 & 0.769 & 0.643 & 0.778 \\
WikiTopicsMT4:sci & 0.283 & 0.448 & 0.278 & 0.447 \\
Metafam & 0.300 & 0.769 & 0.408 & 0.856 \\
FBNELL & 0.464 & 0.644 & 0.485 & 0.647 \\
\bottomrule
\end{tabular}
\end{table}
\begin{table}[!p]
\centering
\caption{Detailed results of the KGFM FLOCK and FLOCK using KMAS as the negative sampling method.}
\label{res:flock_transductive}
\begin{tabular}{ccccc}
\toprule
\multirow{2}{*}{\bf{\textit{Data set}}} & \multicolumn{2}{c}{FLOCK} & \multicolumn{2}{c}{FLOCK + KMAS} \\
\cmidrule(l){2-3} \cmidrule(l){4-5}
& \bf{\textit{MRR}} & \bf{\textit{Hits@10}} & \bf{\textit{MRR}} & \bf{\textit{Hits@10}} \\
\midrule
\multicolumn{5}{c}{\textit{Transductive}} \\
\midrule
CoDExSmall & 0.443 & 0.626 & 0.446 & 0.644 \\
CoDExLarge & 0.342 & 0.465 & 0.342 & 0.467 \\
NELL995 & 0.487 & 0.627 & 0.499 & 0.661 \\
WDsinger & 0.412 & 0.523 & 0.418 & 0.531 \\
NELL23k & 0.217 & 0.372 & 0.227 & 0.381 \\
FB15k237\_10 & 0.245 & 0.402 & 0.246 & 0.402 \\
FB15k237\_20 & 0.274 & 0.445 & 0.275 & 0.448 \\
FB15k237\_50 & 0.320 & 0.520 & 0.325 & 0.524 \\
Hetionet & 0.246 & 0.371 & 0.249 & 0.380 \\
\midrule
\multicolumn{5}{c}{\textit{Inductive (e)}} \\
\midrule
WN18RRInductive:v1 & 0.653 & 0.790 & 0.687 & 0.802 \\
WN18RRInductive:v2 & 0.619 & 0.776 & 0.695 & 0.793 \\
WN18RRInductive:v3 & 0.463 & 0.623 & 0.482 & 0.624 \\
WN18RRInductive:v4 & 0.641 & 0.728 & 0.652 & 0.729 \\
FB15k237Inductive:v1 & 0.486 & 0.680 & 0.498 & 0.696 \\
FB15k237Inductive:v2 & 0.530 & 0.737 & 0.533 & 0.742 \\
FB15k237Inductive:v3 & 0.513 & 0.685 & 0.513 & 0.687 \\
FB15k237Inductive:v4 & 0.505 & 0.703 & 0.502 & 0.701 \\
NELLInductive:v2 & 0.566 & 0.774 & 0.569 & 0.782 \\
NELLInductive:v3 & 0.589 & 0.787 & 0.576 & 0.776 \\
NELLInductive:v4 & 0.559 & 0.786 & 0.547 & 0.786 \\
ILPC2022:small & 0.310 & 0.465 & 0.313 & 0.465 \\
ILPC2022:large & 0.320 & 0.442 & 0.321 & 0.441 \\
HM:3k & 0.066 & 0.127 & 0.064 & 0.113 \\
HM:5k & 0.068 & 0.130 & 0.063 & 0.112 \\
HM:indigo & 0.427 & 0.639 & 0.429 & 0.639 \\
\midrule
\multicolumn{5}{c}{\textit{Inductive (e, r)}} \\
\midrule
FBIngram:100 & 0.450 & 0.660 & 0.456 & 0.660 \\
FBIngram:75 & 0.419 & 0.624 & 0.420 & 0.627 \\
FBIngram:50 & 0.354 & 0.566 & 0.353 & 0.567 \\
FBIngram:25 & 0.405 & 0.666 & 0.408 & 0.665 \\
WKIngram:100 & 0.185 & 0.303 & 0.184 & 0.302 \\
WKIngram:75 & 0.373 & 0.534 & 0.383 & 0.540 \\
WKIngram:50 & 0.129 & 0.269 & 0.131 & 0.274 \\
WKIngram:25 & 0.275 & 0.465 & 0.275 & 0.464 \\
NLIngram:75 & 0.302 & 0.498 & 0.306 & 0.537 \\
NLIngram:50 & 0.329 & 0.510 & 0.339 & 0.534 \\
NLIngram:25 & 0.315 & 0.520 & 0.352 & 0.566 \\
WikiTopicsMT1:tax & 0.281 & 0.394 & 0.279 & 0.368 \\
WikiTopicsMT2:org & 0.101 & 0.167 & 0.102 & 0.167 \\
WikiTopicsMT2:sci & 0.320 & 0.462 & 0.316 & 0.441 \\
WikiTopicsMT3:art & 0.305 & 0.469 & 0.307 & 0.470 \\
WikiTopicsMT3:infra & 0.668 & 0.807 & 0.677 & 0.819 \\
WikiTopicsMT4:sci & 0.302 & 0.473 & 0.303 & 0.473 \\
Metafam & 0.403 & 0.910 & 0.422 & 0.891 \\
FBNELL & 0.514 & 0.707 & 0.493 & 0.685 \\
\bottomrule
\end{tabular}
\end{table}

\end{document}